\documentclass[10pt,twocolumn,letterpaper]{article}

\usepackage[pagenumbers]{cvpr} % To force page numbers, e.g. for an arXiv version
\usepackage{multirow}
\usepackage{booktabs}
\usepackage{arydshln}
\usepackage{float}
\usepackage{url}

\usepackage{amsmath}
\usepackage{amssymb}
\usepackage{pifont}
\usepackage{xcolor}
\definecolor{lightblue}{RGB}{70, 130, 230}
\definecolor{lightgreen}{RGB}{100, 160, 90}
\usepackage[percent]{overpic}
\usepackage{adjustbox}
\usepackage{soul}

\usepackage{lineno}

\definecolor{tableblue}{RGB}{244, 199, 195} 
\definecolor{tablegreen}{RGB}{218, 232, 252}
\definecolor{tablegray}{RGB}{213, 232, 212}

\newcommand{\hlgreen}[1]{\sethlcolor{tablegreen}\hl{#1}}
\newcommand{\hlblue}[1]{\sethlcolor{tableblue}\hl{#1}}
\newcommand{\hlgray}[1]{\sethlcolor{tablegray}\hl{#1}}

\definecolor{cvprblue}{rgb}{0.21,0.49,0.74}
\usepackage[pagebackref,breaklinks,colorlinks,citecolor=cvprblue]{hyperref}

\def\paperID{7358} % *** Enter the Paper ID here
\def\confName{CVPR}
\def\confYear{2025}

\title{MoEdit: On Learning Quantity Perception for Multi-object Image Editing}

\author{Yanfeng Li$^{1}$ \quad Kahou Chan$^{1}$\quad Yue Sun$^{1}$ \quad Chantong Lam$^{1}$ \quad Tong Tong$^{2}$\\ \quad Zitong Yu$^3$ \quad  Keren Fu$^4$ \quad Xiaohong Liu$^{5*}$ \quad Tao Tan$^{1}\thanks{~Corresponding author.}$\\
	$^{1}$ Macao Polytechnic University \quad $^{2}$ Fuzhou University \quad $^{3}$ Great Bay University \\ \quad $^{4}$ Sichuan University \quad $^{5}$ Shanghai Jiao Tong University
}

\begin{document}

\maketitle

\begin{abstract}
Multi-object images are prevalent in various real-world scenarios, including augmented reality, advertisement design, and medical imaging. Efficient and precise editing of these images is critical for these applications. With the advent of Stable Diffusion (SD), high-quality image generation and editing have entered a new era. However, existing methods often struggle to consider each object both individually and part of the whole image editing, both of which are crucial for ensuring consistent quantity perception, resulting in suboptimal perceptual performance.
To address these challenges, we propose MoEdit, an auxiliary-free multi-object image editing framework. MoEdit facilitates high-quality multi-object image editing in terms of style transfer, object reinvention, and background regeneration, while ensuring consistent quantity perception between inputs and outputs, even with a large number of objects. To achieve this, we introduce the \underline{Fe}ature \underline{Com}pensation (FeCom) module, which ensures the distinction and separability of each object attribute by minimizing the in-between interlacing. Additionally, we present the \underline{Q}uantity A\underline{t}\underline{t}entio\underline{n} (QTTN) module, which perceives and preserves quantity consistency by effective control in editing, without relying on auxiliary tools.
By leveraging the SD model, MoEdit enables customized preservation and modification of specific concepts in inputs with high quality. Experimental results demonstrate that our MoEdit achieves State-Of-The-Art (SOTA) performance in multi-object image editing. Data and codes will be available at \url{https://github.com/Tear-kitty/MoEdit}.
\end{abstract}

\section{Introduction}
\label{sec:intro}

Multi-object image editing enables flexible modification of various concepts in images while maintaining overall visual coherence \cite{chakrabarty2024lomoe}. These techniques are essential for applications such as image generation \cite{sheynin2024emu, matsuda2024multi}, medical imaging \cite{shi2024dragdiffusion, wang2024diffusion}, and augmented reality \cite{cao2023mobile, martin2023sttar, li2024gbot}.

Despite its importance, editing multi-object images remains a challenging task due to existing methods struggling to consider each object both individually and part of the whole image. This often results in inconsistent quantity perception during editing, leading to visual distortions, blurring, and disruptions in scene semantics and logical structure, as well as reduced editability \cite{wang2023enhancing, liu2024structure}.

Traditional techniques typically focus on enhancing single-object editing by fine-tuning methods such as Stable Diffusion (SD) locally, but fail to ensure consistent quantity perception across multiple objects \cite{shi2024dragdiffusion, mou2024diffeditor, joseph2024iterative}. Recent approaches leveraging Large Language Models (LLMs) for text-image alignment have focused on maintaining consistency in the global information of the whole image. While these methods ensure quantity consistency visually \cite{wu2024self, xu2024inversion}, they often encounter aliasing of object attributes.
%, which arises from neglecting to consider each object individually. 
This oversight makes the effective extraction of object attributes difficult during editing \cite{mokady2023null, titov2024guide}.
Auxiliary tools, such as masks, are commonly used to extract object attributes \cite{avrahami2023blended, couairon2022diffedit}. While effective, this approach fails to capture global image information of the whole image and requires a one-to-one association between auxiliary tools and each object to preserve quantity consistency. This significantly increases training costs, reduces user convenience, and often demands manual adjustments. Therefore, there is an urgent need for a solution that extracts object attributes from the whole image, ensuring the distinction and separability of these attributes while simultaneously perceiving global information of the whole image. Such a solution would eliminate the need for consistent auxiliary tool quantities and address limitations in the multi-object image editing.

To address this technical challenge, we propose the MoEdit pipeline, an auxiliary-free solution designed to enhance the editing quality and editability of multi-object images by ensuring consistent quantity perception between inputs and outputs. Our approach comprises two key modules: 
(1) Feature Compensation (FeCom) module: A feature attention is employed to enable interaction between text prompts, which contain quantity and object information, and the inferior image features extracted by the image encoder of CLIP \cite{radford2021learning}. This process compensates for inferior features to minimize the in-between interlacing, and ultimately extracts object attributes. The enhanced attribute distinction of each object also provides a foundation for the QTTN module to achieve auxiliary-free quantity perception. 
(2) Quantity Attention (QTTN) module: Without relying on any auxiliary tools, this module perceives the information of each object both individually and part of the whole image from the image features enhanced by FeCom. The perceived information is then injected into specific blocks of U-Net \cite{podell2023sdxl}, enabling the model to acquire a clear and consistent quantity perception. The visual comparisons of our MoEdit with the latest method TurboEdit \cite{wu2024turboedit} is shown in Fig.~\ref{homepage}.

Our contributions are as follows:
\begin{itemize}
	\item We propose a novel pipeline named MoEdit to pioneer consistent quantity perception for multi-object image editing, even with a large number of objects. This design provides inspiring insights for such challenges. 
	\item We introduce the FeCom and QTTN modules to extract object attributes and preserve quantity consistency without auxiliary tools, respectively.
	\item Compared to existing methods, MoEdit shows the SOTA performance in the preservation of quantity consistency, visual quality and editability.
\end{itemize}

\vspace{-3.7pt}
\section{Related Works}
\subsection{Text-to-Image Diffusion Models}
Diffusion models have emerged as a groundbreaking research direction. These models are widely utilized in video generation \cite{liu2025improving, li2024q, yang2024diffstega, li2024g}, QR code generation \cite{wu2024text2qr, cui2025face2qr}, and audio synthesis \cite{lee2023imaginary, choi2024dddm, xue2024singvisio}. Among these applications, Text-to-Image diffusion (T2I) models, which generate images from user-provided textual prompts, have garnered significant attention \cite{croitoru2023diffusion, zhao2024uni, li2024snapfusion}. Furthermore, the generated images also provide reliable data sources for advancing the field of image quality assessment \cite{zhang2024bench, li2023agiqa, zhang2023perceptual}. 

\subsection{Image Editing}
The application of diffusion models to image editing 
%began with SDEdit \cite{meng2021sdedit}, which introduced new perspectives in this domain. Subsequent advancements 
can be categorized into three primary approaches:
(1) fine-tuning approaches: Methods like Imagic \cite{kawar2023imagic} fine-tune diffusion models using the original image or target prompts \cite{zhang2023sine, wang2024instancediffusion}. While effective for specific tasks, these methods face reduced efficiency. (2) internal representation approaches: This line of research leverages internal model representations to address efficiency bottlenecks without requiring fine-tuning \cite{tumanyan2023plug, xu2023inversion, ling2024freedrag}. 
%For instance, interactive batch editing techniques utilize these representations to edit image sets \cite{nguyen2024edit}. 
However, these methods often struggle to maintain the original structure and details in certain tasks. (3) inversion optimization approaches: A third category focuses on enhancing inversion processes to minimize errors during generation \cite{wallace2023edict, huberman2024edit}. Techniques like WaveOpt-Estimator \cite{koo2024wavelet} and Null-text Inversion \cite{mokady2023null} improve detail preservation and structural stability while significantly enhancing editing efficiency. Despite their advantages, these methods only target single-object attribute. %or lack refined control for complex scenarios.

\begin{table}[h]
	\centering
	\renewcommand{\arraystretch}{1.1}
	\LARGE
	\resizebox{\linewidth}{!}{%
		\begin{tabular}{cccccc}
			\hline
			Method      & w/o LLM & w/o Guidence & Base Model & Quantity Consistency & Object Attributes \\ \hline
			SSR-Encoder \cite{zhang2024ssr} & \ding{51}       & \ding{55}            & SD1.5          & \ding{55}      & \ding{51}          \\
			$\lambda$-Eclipse \cite{patel2024lambda}  & \ding{51}       & \ding{51}            & Kandinsky          & \ding{55}      & \ding{55}          \\
			IP-Adapter \cite{ye2023ip}          & \ding{51}       & \ding{51}            & SDXL          & \ding{55}      & \ding{55}          \\
			Blip-diffusion \cite{li2024blip}        & \ding{55}       & \ding{51}            & SD1.5          & \ding{55}      & \ding{55}          \\
			MS-diffusion \cite{wang2024ms}          & \ding{51}       & \ding{55}            & SDXL          & \ding{55}      & \ding{51}          \\
			Emu2 \cite{sun2024generative}        & \ding{55}       & \ding{51}            & SDXL          & \ding{51}      & \ding{55}          \\
			TurboEdit  \cite{wu2024turboedit}         & \ding{55}       & \ding{51}            & SDXL          & \ding{51}      & \ding{55}          \\
			MoEdit (ours)          & \ding{51}       & \ding{51}            & SDXL          & \ding{51}      & \ding{51}          \\ \hline
		\end{tabular}%
	}
	\caption{An overview of previous studies. “w/o LLM” and “w/o Guidance” indicate whether the model utilizes an LLM or specific guidance as auxiliaries, respectively. Object Attributes and Quantity Consistency denote whether the method considers each object individually or part of the whole image, respectively.}
	\label{tab:comparison attribute}
\end{table}
\subsection{Quantity Perception in Image Generation}
Consistent quantity perception requires models to consider each object both individually and part of the whole image. This capability allows a model to understand the interplay between quantity, semantics, and image structure. Some approaches leverage LLMs to align textual and visual elements for finer editing \cite{wu2024self}. However, they often neglect to ensure the distinction and separability of each object attribute \cite{huang2024smartedit, wang2024genartist, li2024blip}. Methods like SLD \cite{zhang2024sgedit}, TurboEdit \cite{wu2024turboedit}, and Emu2 \cite{sun2024generative} achieve visual consistency in quantity but are constrained to editing a single object in multi-object scenarios, limiting their broader applicability. 
In contrast, object-centered approaches such as LoMOE \cite{chakrabarty2024lomoe}, MS-diffusion \cite{wang2024ms}, and SSR-Encoder \cite{zhang2024ssr} use masks, bounding boxes, or dedicated target queries to extract individual object attributes. While effective, these methods face challenges in modeling each object as part of the whole image, necessitating one-to-one associations with auxiliary tools to preserve quantity consistency. However, this reliance on auxiliary tools increases both training complexity and cost, reducing user convenience. In this work, we propose a pipeline that can effectively extract object attributes and preserve quantity consistency without auxiliaries (LLM and Guidance). It simply and efficiently implements quantity perception for multi-object images. Some previous methods are reported in Table~\ref{tab:comparison attribute}.

\begin{figure}[t]
	\centering
	\begin{overpic}[width=0.5\textwidth, trim=28 20 10 15, clip]{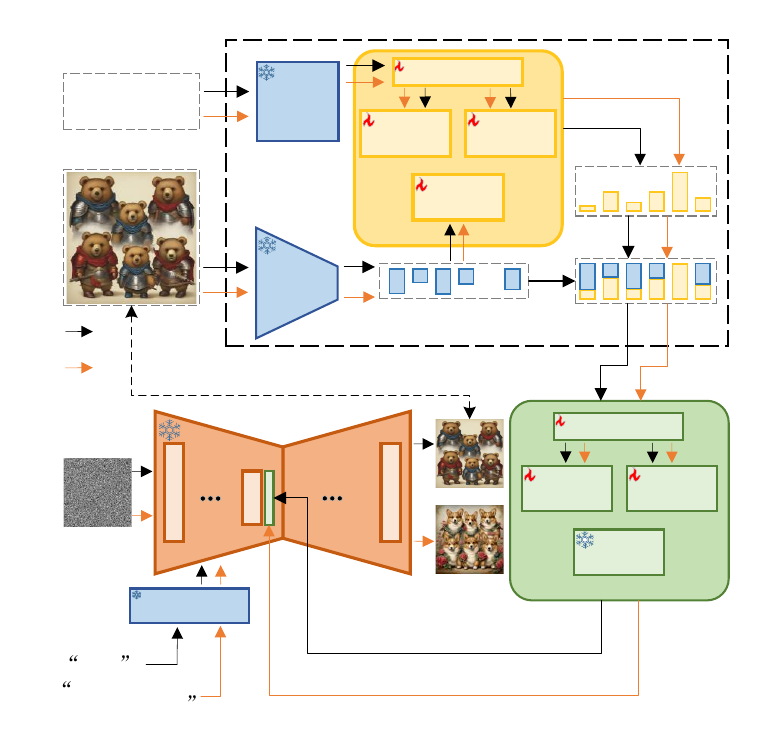}
		\put(9,94){\footnotesize\parbox{2cm}{\textcolor{lightblue}{{$c_q$}}}}
		\put(1.8,87.5){\footnotesize\parbox{2cm}{\textcolor{lightblue}{{\textit{six doll bears}}}}}
		\put(0.8,2){\scriptsize\parbox{1.8cm}{\centering\textit{corgis, roses in background}}}
		\put(4,13){\scriptsize\parbox{2cm}{$c_e$}}
		\put(1.5,10){\scriptsize\parbox{2cm}{\textit{null-text}}}
		\put(4,7){\footnotesize\parbox{2cm}{\textit{ $\varnothing$ }}}
		\put(-0.5,51.5){\scriptsize\parbox{2cm}{{Training}}}
		\put(-0.5,47){\scriptsize\parbox{2cm}{{Inference}}}
		\put(36,68.8){\footnotesize\parbox{4cm}{\centering{Feature\ \ \ \ \ Attention}}}
		\put(64,92.3){\footnotesize\parbox{4cm}{\centering{Feature\\ Compensation}}}
		\put(23,62.2){\footnotesize\parbox{2cm}{\centering{Image\\ Encoder}}}
		\put(23.4,87.7){\footnotesize\parbox{2cm}{\centering{Text\\ Encoder}}}
		\put(46.5,92.2){\scriptsize\parbox{2cm}{\centering{FC}}}
		\put(54.2,83.2){\scriptsize\parbox{2cm}{\centering{V\\ Projection}}}
		\put(39,83.2){\scriptsize\parbox{2cm}{\centering{K\\ Projection}}}
		\put(46.5,74){\scriptsize\parbox{2cm}{\centering{Q\\ Projection}}}
		\put(12,47){\footnotesize\parbox{4cm}{\centering{Reconstruction Loss}}}
		\put(62.3,31.7){\scriptsize\parbox{2cm}{\centering{K\\ Projection}}}
		\put(77.5,31.7){\scriptsize\parbox{2cm}{\centering{V\\ Projection}}}
		\put(70,22.4){\scriptsize\parbox{2cm}{\centering{Q\\ Projection}}}
		\put(70,40.7){\scriptsize\parbox{2cm}{\centering{Extraction}}}
		\put(70,17){\scriptsize\parbox{2cm}{\centering{Quantity Attention}}}
		\put(8.8,14.5){\scriptsize\parbox{1.8cm}{\centering{Text Encoder}}}
		%\put(48,5.5){\footnotesize\parbox{1.8cm}{\centering{Insert}}}
	\end{overpic}
	\vspace{-12pt}
	\caption{{The framework of our propose method}. The Feature Compensation (FeCom) module uses text prompts with quantity and object information to compensate for the inferior image features extracted by the image encoder of the CLIP. The compensated image features are then processed by the Quantity Attention (QTTN) module to conduct consistent quantity perception, which are injected into the U-Net to control image editing. During training, the text prompts is set to null-text. During inference, the text prompts can be modified to edit images.}
	\label{algorithm_framework}
\end{figure}

\begin{figure*}[t]
	\centering
	\begin{overpic}[width=1\textwidth, trim=20 20 15 15, clip]{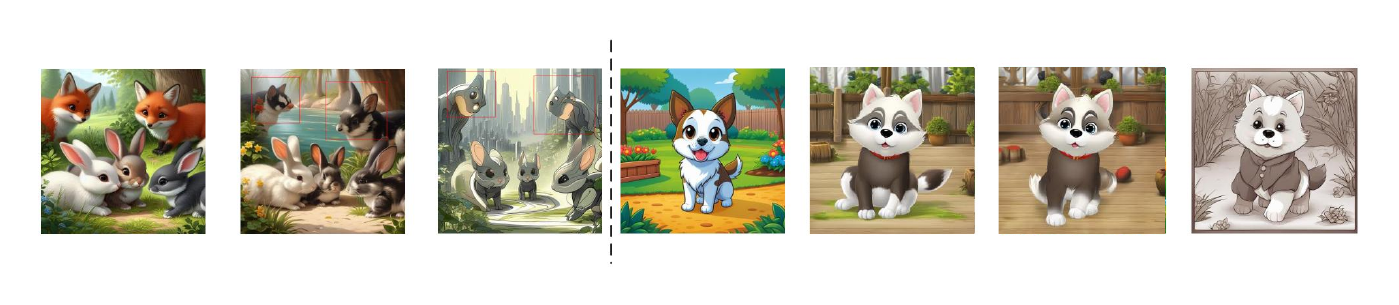}
		\put(2.5,15.5){\small \parbox{2cm}{{{Reference}}}}
		\put(46.2,15.5){\small \parbox{2cm}{{{Reference}}}}
		\put(26.2,15.5){\small \parbox{2cm}{{{Output}}}}
		\put(76,15.5){\small \parbox{2cm}{{{Output}}}}
		\put(15.7,-0.5){\footnotesize \parbox{4cm}{\textit{“...by the pool”}}}
		\put(30,-0.5){\footnotesize \parbox{2cm}{\centering\textit{“...futuristic city style”}}}
		\put(58,-0.5){\footnotesize \parbox{2cm}{\centering\textit{\textcolor{red}{+}$\varnothing$}}}
		\put(67.2,-0.5){\footnotesize \parbox{4cm}{\centering\textit{\textcolor{red}{+} $random\ \mathcal{N}(0,1)$}}}
		\put(81.8,-0.5){\footnotesize \parbox{4cm}{\centering\textit{\textcolor{red}{+} $random\ \mathcal{N}(0,1)$}}}
		\put(0,-0.5){\small \parbox{2cm}{{{(a)}}}}
		\put(43.7,-0.5){\small \parbox{2cm}{{{(b)}}}}
		%\put(24,12){\parbox{2cm}{\tiny \rotatebox{90}{{$B_4$}}}}
	\end{overpic}
	\vspace{-10pt}
	\caption{{The illustration of the purpose of FeCom module}. Reference denotes the input image. (a) When only the CLIP-encoded image information is used to represent $I_g$ as input to the QTTN module, the attributes of foxes are either lost or shifted towards the attributes of rabbits, resulting in attributes aliasing. (b) The three images illustrate the results of adding three different Gaussian noises, $\mathcal{N}(0,1)$, to $CLIP(I)$. This process alters the object attributes, background, and style while maintaining the structure and clarity of the image.}
	\label{observation_cd}
\end{figure*}
\section{Method}
\subsection{Overall}
We propose MoEdit, denoted as $\mathcal{G}$, with its architecture illustrated in Fig.~\ref{algorithm_framework}. MoEdit builds upon the foundation of the SD model, extending its capabilities to perceive the information of each object both individually and part of the whole image within input image $I$. This ensures a clear understanding of quantity perception while preserving attributes of each object. 

The primary objective of MoEdit is to preserve quantity consistency in multi-object image editing, thereby delivering high-quality results that are more closely aligned with user expectations. The edited output image $I_e = \mathcal{G}(c_e, c_q, I, z_0)$ is generated based on two textual conditions $c_e$ and $c_q$, an image condition $I$, and noise $z_0$. Specifically, $c_e$ denotes the editing instruction, while $c_q$ serves as a caption specifying the names and quantities of objects present in the input image. The architecture achieves these objectives through two modules: the FeCom and QTTN module.

The FeCom module, denoted as $\mathcal{F}$, processes the user-provided inputs $I$ and $c_q$. This module bridges user-defined text prompts, which contain quantity and object information, with the inferior CLIP-encoded features extracted from the input image $I$, minimizing in-between interlacing to achieve the extraction of object attributes. The output of the FeCom module is a set of enhanced image features $I_g$, which ensure the distinction and separability of each object attribute, serving as input for the subsequent QTTN module.

The QTTN module, denoted as $\mathcal{Q}$, considers to perceive the information of each object both individually and part of the whole context within $I_g$. By injecting this information into a specific block of the U-Net, QTTN enables to preserve quantity consistency by effective control in editing process, without relying on auxiliary tools. This ensures efficient editing performance and high editability.
%For a detailed design, please refer to {Supplementary Material}.

During the training phase, $c_e$ is set as a null-text input to isolate the learning of consistent quantity perception from textual embedding interference. This design allows the FeCom and QTTN modules to focus on their respective tasks. At inference, $c_e$ can be customized by users to guide the generation of high-quality images that meet their specific expectations. The subsequent sections detail the design and implementation of the FeCom and QTTN modules, along with the training and inference stages.

\subsection{Feature Compensation}
Relying solely on image features encoded by CLIP to represent $I_g$ as input to the QTTN module often leads to aliasing of object attributes, as shown in Fig.~\ref{observation_cd} (a). This issue stems from strength of CLIP in capturing general visual semantics, yet its difficulty in encoding distinct, separable features in multi-object images. To mitigate this, the FeCom module leverages user-provided text prompts, which contain object and quantity information, to compensate the inferior image features encoded by CLIP. Through binding object and quantity information to minimizing the in-between interlacing between multiple objects, the FeCom module enhances the distinction and separability of each object attribute, thereby extracting object attributes and supporting high-quality editing.
\paragraph{Object Preservation.} 
The FeCom module consists of three key components, as illustrated in Fig.~\ref{algorithm_framework}, a CLIP-based image encoder $CLIP(\cdot)$, a feature attention mechanism, and a feature compensation mechanism. First, the input image $I$ is encoded into $CLIP(I)$, while the text prompt $c_q$ serves as input to the attention mechanism. The feature attention mechanism maps object-specific quantity information from the text prompt onto $CLIP(I)$, identifying gaps in the encoded image features. This process generates compensation features that align with the original image features. Mathematically, this interaction is expressed as:
\begin{equation}
	\begin{split}
		I_c=Softmax(\frac{Q_gK_t^T}{\sqrt{d_k}})V_t,\qquad\qquad\qquad\quad\ \ \ \ \ \\
		K_t=F(c_q),\ V_t=F(c_q),\ Q_g=F(CLIP(I)),
	\end{split}
	\label{attention_FeCom}
\end{equation}
with $F(\cdot)$ representing an MLP layer. The resulting compensation features $I_c$ are then added to $CLIP(I)$ to produce enhanced image features $I_g$, ensuring the distinction and separability of each object attribute. This process is formulated as:
\begin{equation}
	\begin{split}
		\mathcal{F}(I, c_q)=I_g=CLIP(I)+\lambda I_c,
	\end{split}
	\label{injcetion_FeCom}
\end{equation}
where $\lambda$ is a tunable scaling factor.

The motivation for this approach, as illustrated in Fig.~\ref{observation_cd} (b), even when Gaussian noise $\mathcal{N}(0, 1)$ is added to the image features extracted by CLIP and used as $I_g$ in the QTTN module, the resulting images do not degrade into incomprehensible or low-quality outputs. Instead, the added noise primarily alters object attributes, background, and style while preserving the structure and clarity of the image. This observation highlights that the source of aliasing of object attributes lies in limited ability of CLIP to minimize the in-between interlacing and encode attribute for each object.

Through this design, the FeCom module ensures the distinction and separability of each object attribute, enabling the model to extract object attributes during editing. Moreover, it enhances the handling of fine image details, ensuring that MoEdit consistently generates high-quality images.

\begin{figure}[t]
	\centering
	\begin{overpic}[width=0.5\textwidth, trim=18 20 10 15, clip]{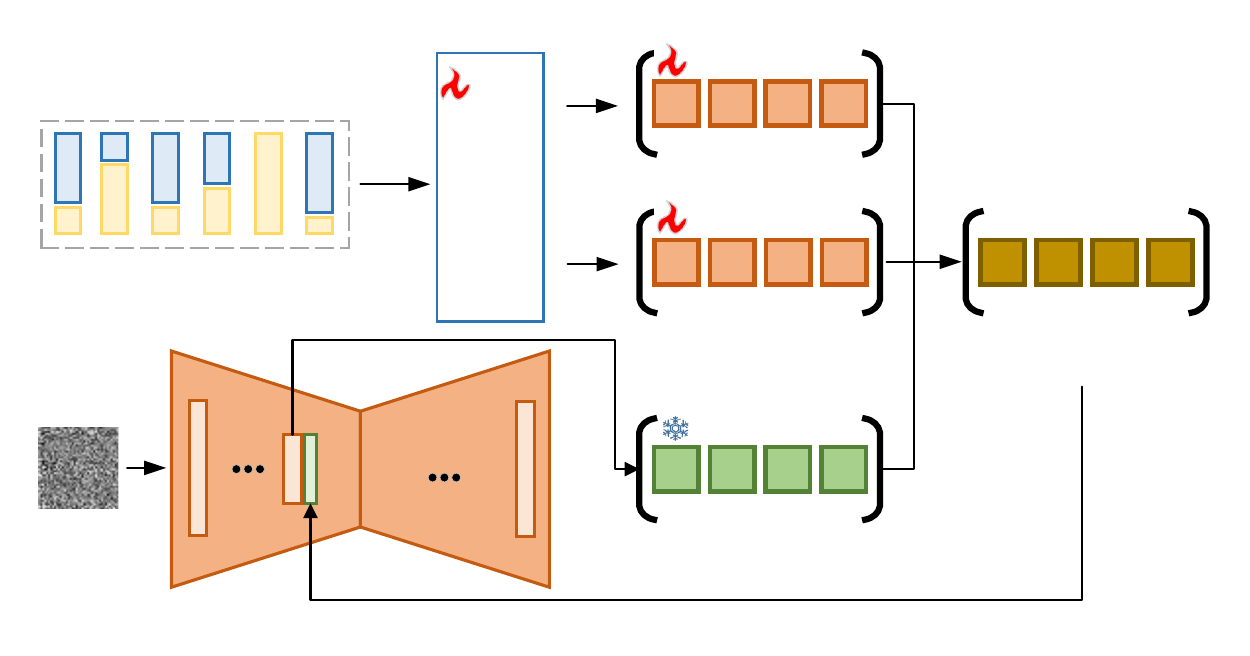}
		\put(36.5,35){\parbox{2cm}{\rotatebox{90}{{Extraction}}}}
		\put(58.5,6.5){\parbox{2cm}{{\textit{$Q_z$}}}}
		\put(58.5,37){\parbox{2cm}{{\textit{$K_g$}}}}
		\put(58.5,23.5){\parbox{2cm}{{\textit{$V_g$}}}}
		\put(83.3,23.5){\parbox{2cm}{{\textit{$V_{new}$}}}}
		%\put(24,12){\parbox{2cm}{\tiny \rotatebox{90}{{$B_4$}}}}
	\end{overpic}
	\vspace{-10pt}
	\caption{{The framework of QTTN module}. The Extraction module perceives the information of each object both individually and part of the whole image. Subsequently, an attention mechanism is employed to convert this information into a format compatible with the U-Net architecture, thereby ensuring that the consistent quantity perception provided by the QTTN module effectively controls the editing process.}
	\label{QTTN}
	\vspace{-5pt}
\end{figure}
\subsection{Quantity Attention}
This module performs two key tasks. First, it treats an image as a unified whole to extract global information, enabling consistent quantity perception essential for quantity preservation. Second, it defines the attribute boundary of each object from enhanced image features $I_g$, which is crucial for high editability. Notably, the module relies solely on internal feature interactions to address these tasks, eliminating the need for auxiliary tools. This approach significantly reduces training costs and enhances user convenience.
\paragraph{Quantity-Perception Control.}
The QTTN module consists of three components, as illustrated in Fig.~\ref{QTTN}, the extraction module $E_t$, attention interaction, and U-Net injection. First, the enhanced image features $I_g$ refined by the FeCom module, and the input noise $z_t^4$ from the fourth block $B_4$ of the U-Net are obtained. Next, $E_t$ extracts the information of each object both individually and part of the whole image from the mixed context $I_g$, which are then passed to the attention interaction stage to interact with $z_t^4$. This process ensures that the format of quantity perception generated by QTTN module is compatible with the U-Net architecture. Mathematically, this is represented as:
\begin{equation}
	\begin{split}
		\mathcal{Q}(I_g, z_t^4)=V_{new}=Softmax(\frac{Q_zK_g^T}{\sqrt{d_k}})V_g,\quad\ \ \ \ \ \ \\
		K_g=F(E_t(I_g)),\ V_g=F(E_t(I_g)),\ Q_z=F(z_t^4),
	\end{split}
	\label{attention_QTTN}
\end{equation}
where $F(\cdot)$ denotes an MLP layer. The resulting $V_{new}$ is directly injected into $B_4$ of the U-Net to control the editing process to ensure consistent quantity perception. The injection is defined as: 
\begin{equation}
	\begin{split}
		z_t^5=Attn(Q_z, K_i, V_i) + \beta V_{new},
	\end{split}
	\label{injcetion_QTTN}
\end{equation}
where $z_t^5$ is the final output of $B_4$ and input to the fifth block $B_5$ of the Unet. $Attn(Q_z, K_i, V_i)$ represents the cross-attention for text promtps $c_e$, $\beta$ is a tunable scaling factor.
\paragraph{Insertion Point.} 
Following prior studies \cite{wang2024instantstyle}, the U-Net is decomposed into 11 transformer blocks, denoted as $B_i$ $(i=1,2,\dots,11)$. These blocks include four downsampling blocks, one middle block, and six upsampling blocks. The fourth block $B_4$ is chosen as the insertion point for the QTTN module, based on the functionality of each block \cite{frenkel2024implicit}. As demonstrated in Sec.~\ref{sec:ablation study}, the insertion point of the $B_4$ provides an optimal balance, allowing the U-Net to maximize the utility of consistent quantity perception while maintaining high editability.

This design empowers the QTTN module to enrich the U-Net model with consistent quantity perception, enabling effective control over object quantities and ensuring high editability during image editing.
\subsection{Training and Inference}
During the training phase, only MSE Loss was employed to guide the learning process. As shown in Fig.~\ref{algorithm_framework}, user-provided text prompts for guiding image editing are replaced with null-text inputs. This prevents interference while retaining their utility for image editing. In the inference phase, users can input personalized text prompts to direct image editing toward their desired outputs.
\section{Experiments}
\begin{table*}[]
	\centering
	\renewcommand{\arraystretch}{1.2}
	\resizebox{\textwidth}{!}{%
		\begin{tabular}{cccccccccccccccccccc}
			\hline
			\multicolumn{2}{c}{\multirow{3}{*}{\textbf{Method}}} & \multicolumn{8}{c}{Objective Metrics}                                                                                                                                                                                        & \multirow{11}{*}{} & \multicolumn{9}{c}{Subjective Metrics}                                                                                                 \\ \cline{3-10} \cline{12-20} 
			\multicolumn{2}{c}{}                                 & \multirow{2}{*}{\textbf{NIQE $\downarrow$}} & \multirow{2}{*}{\textbf{HyperIQA $\uparrow$}} & \multicolumn{2}{c}{\textbf{CLIP Score}} & \multirow{2}{*}{\textbf{LPIPS $\downarrow$}} & \multicolumn{2}{c}{\textbf{Q-Align}}  & \multirow{2}{*}{\textbf{AesBench $\uparrow$}} &                    & \multirow{2}{*}{\textbf{MOS $\uparrow$}} & \multicolumn{8}{c}{\textbf{Numerical $\uparrow$}}                                                                 \\ \cline{5-6} \cline{8-9} \cline{13-20} 
			\multicolumn{2}{c}{}                                 &                                &                                    & \textbf{Whole $\uparrow$}      & \textbf{Edit $\uparrow$}     &                                 & \textbf{Quality $\uparrow$} & \textbf{Aesthetic $\uparrow$} &                                    &                    &                               & \textbf{3} & \textbf{4} & \textbf{5} & \textbf{6} & \textbf{7} & \textbf{8} & \textbf{9} & \textbf{9+} \\ \cline{1-10} \cline{12-20} 
			SSR-Encoder                     & \cite{zhang2024ssr}                  & 3.106                              & 62.16                                  & 0.2897                   & 0.1897                 & 0.2951                               & 4.1175                & 3.7273                  & 67.75                                  &                    & 0.68                             & 64.57          & 59.85          & 21.99          & 25.87          & 14.21          & 5.41          & 6.21          & 3.75           \\
			lambda-Eclipse                  & \cite{patel2024lambda}                  & 3.552                              & 65.42                                  & 0.2963                   & 0.1821                 & 0.3261                               & 4.3247                & 4.1578                  & 68.08                                  &                    & 0.00                             & 82.32          & 83.01          & 50.21          & 55.42          & 41.37          & 35.46          & 40.25          & 25.12           \\
			IP-Adapter                      & \cite{ye2023ip}                  & 2.952                              & \hlgray{71.19}                                  & 0.2938                   & 0.2218                 & 0.3397                               & 4.5724                & 4.2302                  & 70.55                                  &                    & 0.98                             & 25.61          & 30.12          & 17.36          & 15.87          & 5.32          & 1.77          & 2.31          & 0.54           \\
			Blip-diffusion                  & \cite{li2024blip}                  & 2.983                              & 70.55                                  & 0.2913                   & 0.2170                 & 0.3797                               & 4.3667                & 3.8476                  & 65.16                                  &                    & 0.00                             & 52.26          & 30.18          & 10.77          & 4.93          & 3.27          & 5.01          & 0.77          & 0.91           \\
			MS-diffusion                    & \cite{wang2024ms}                  & \hlgreen{2.771}                              & 65.41                                  & \hlgray{0.3043}                   & \hlgray{0.2361}                 & 0.2904                               & \hlgray{4.7645}                & \hlgreen{4.3633}                  & \hlgray{74.45}                                  &                    & \hlgray{7.68}                             & 74.33          & 51.26          & 38.21          & 21.36          & 24.32          & 27.05          & 13.2          & 5.68           \\
			Emu2                            & \cite{sun2024generative}                  & 3.059                              & 68.51                                  & 0.3012                   & 0.2296                 & \hlgreen{0.1875}                               & \hlgreen{4.7836}                & \hlgray{4.3528}                  & \hlgreen{75.09}                                  &                    & 3.87                             & \hlgreen{92.82}          & \hlblue{93.04}          & \hlblue{88.76}          & \hlgreen{82.56}          & \hlblue{84.57}          & \hlblue{80.75}          &\hlblue{82.97}           & \hlblue{73.46}           \\
			TurboEdit                       & \cite{wu2024turboedit}                  & \hlgray{2.872}                              & \hlgreen{71.72}                                  & \hlgreen{0.3101}                   & \hlgreen{0.2427}                 & \hlblue{0.1684}                               & 4.6849                & 4.2181                  & 72.13                                  &                    & \hlgreen{11.35}                             & \hlgreen{93.71}          & \hlgray{91.98}          & \hlgreen{85.78}         &  \hlgray{78.65}         & \hlgray{73.21}          & \hlgray{74.78}          & \hlgray{69.38}          & \hlgray{59.33}           \\
			MoEdit(Ours)                    &                   & \hlblue{2.749}                              & \hlblue{75.66}                                  & \hlblue{0.3254}                   & \hlblue{0.2557}                 & \hlgray{0.2555}                               & \hlblue{4.8763}                & \hlblue{4.6172}                  & \hlblue{77.05}                                  &                    & \hlblue{75.44}                             & \hlblue{95.54}          & \hlgreen{92.15}          & \hlgray{83.44}          & \hlblue{84.31}          & \hlgreen{82.56}          & \hlgreen{77.98}          & \hlgreen{79.65}          & \hlgreen{70.34}           \\ \hline
		\end{tabular}%
	}\vspace{3pt}
	\caption{{Quantitative comparisons on six objective metrics and two subjective metrics}. The \protect\hlblue{best}, \protect\hlgreen{second best} and \protect\hlgray{third best} have been marked by red, blue and green respectively. MoEdit demonstrates superior performance across all objective metrics except LPIPS. In subjective evaluations, while Emu2 slightly outperformed MoEdit in Numerical Accuracy, MoEdit achieves a significant advantage in MOS. In particular, MoEdit produces high-quality edits that effectively preserve quantity consistency and extract object attributes, meeting user expectations and aesthetic preferences.}
	\label{tab:objective metrics}
\end{table*}
\begin{figure*}[t]
	\centering
	\begin{overpic}[width=1\textwidth, trim=17 20 17 15, clip]{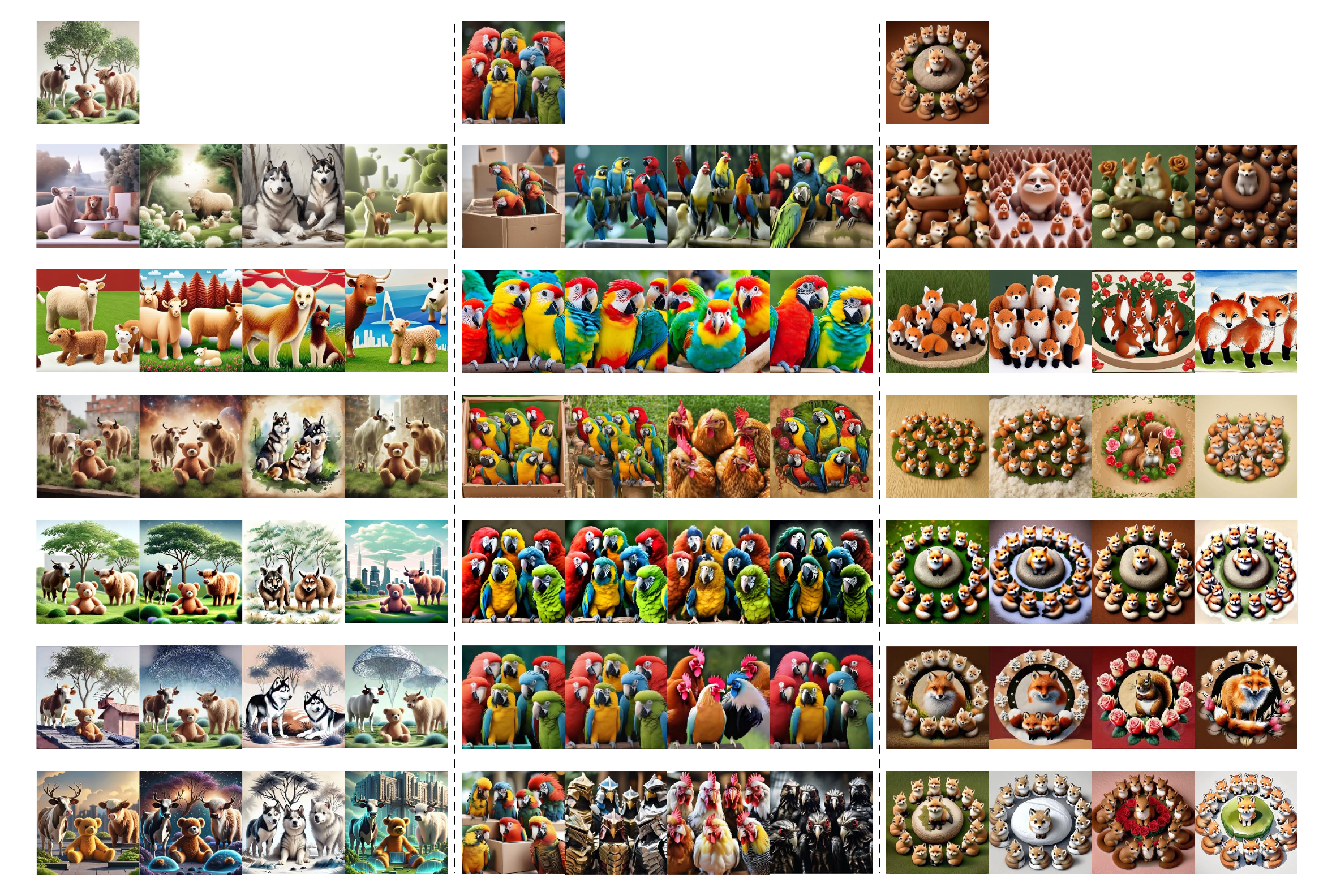}
		\put(0.8,-3.2){\scriptsize \parbox{1.5cm}{\centering\textit{“...on the rooftop”}}}
		\put(9.8,-3.2){\scriptsize \parbox{1.2cm}{\centering\textit{“...in the universe”}}}
		\put(17,-3.2){\scriptsize \parbox{1.5cm}{\centering\textit{“\textcolor{red}{$\rightarrow$}huskies, ink painting style”}}}
		\put(26,-3.2){\scriptsize \parbox{1.1cm}{\centering\textit{“...futuristic city style”}}}
		\put(35.5,-3.2){\scriptsize \parbox{1cm}{\centering\textit{“...inside the box”}}}
		\put(43.5,-3.2){\scriptsize \parbox{1.1cm}{\centering\textit{“...with armors”}}}
		\put(50.6,-3.2){\scriptsize \parbox{1.2cm}{\centering\textit{“\textcolor{red}{$\rightarrow$}chickens”}}}
		\put(58.3,-3.2){\scriptsize \parbox{1.5cm}{\centering\textit{“...dark horror style”}}}
		\put(67.2,-3.2){\scriptsize \parbox{1.5cm}{\centering\textit{“...in a natural field”}}}
		\put(76.3,-3.2){\scriptsize \parbox{1.2cm}{\centering\textit{“...in the snow”}}}
		\put(83.5,-3.2){\scriptsize \parbox{1.5cm}{\centering\textit{“\textcolor{red}{$\rightarrow$}squirrels, rose in background”}}}
		\put(92.1,-3.2){\scriptsize \parbox{1.2cm}{\centering{“...watercolor style”}}}
		\put(0,58.7){\scriptsize \rotatebox{90}{\parbox{1.2cm}{\centering{Reference}}}}
		\put(0,49.2){\scriptsize \rotatebox{90}{\parbox{1.2cm}{\centering{IP-Adapter}}}}
		\put(0,37.4){\scriptsize \rotatebox{90}{\parbox{2cm}{\centering{Blip-diffusion}}}}
		\put(0,27.5){\scriptsize \rotatebox{90}{\parbox{2cm}{\centering{MS-diffusion}}}}
		\put(0,19.5){\scriptsize \rotatebox{90}{\parbox{1.2cm}{\centering{Emu2}}}}
		\put(0,9.8){\scriptsize \rotatebox{90}{\parbox{1.2cm}{\centering{TurboEdit}}}}
		\put(0,-0.1){\scriptsize \rotatebox{90}{\parbox{1.2cm}{\centering{MoEdit(Ours)}}}}
		\put(9.5,61.5){\small\parbox{4cm}{\centering\textcolor{lightblue}{\textit{two cows and a teddy bear}}}}
		\put(43,61.5){\small\parbox{4cm}{\centering\textcolor{lightblue}{\textit{seven parrots}}}}
		\put(76.5,61.5){\small\parbox{4cm}{\centering\textcolor{lightblue}{\textit{fourteen foxes}}}}
	\end{overpic}
	\vspace{8pt}
	\caption{{Qualitative comparisons}. Blip-diffusion \cite{li2024blip} struggles with editability and stability, while IP-Adapter \cite{ye2023ip} improves clarity but lacks robustness in quantity preservation. MS-diffusion \cite{wang2024ms} handles few objects well but fails with larger numbers. Emu2 \cite{sun2024generative} ensures structural consistency but limits editability. TurboEdit \cite{wu2024turboedit} balances tasks yet falls short in quantity and style handling. Finally, MoEdit successfully achieves a well-balanced performance in quantity consistency, semantic accuracy, stylistic consistency, and aesthetic quality.}
	\label{comparisons}
\end{figure*}
\begin{figure*}[t]
	\centering
	\begin{overpic}[width=1\textwidth, trim=18 10 18 15, clip]{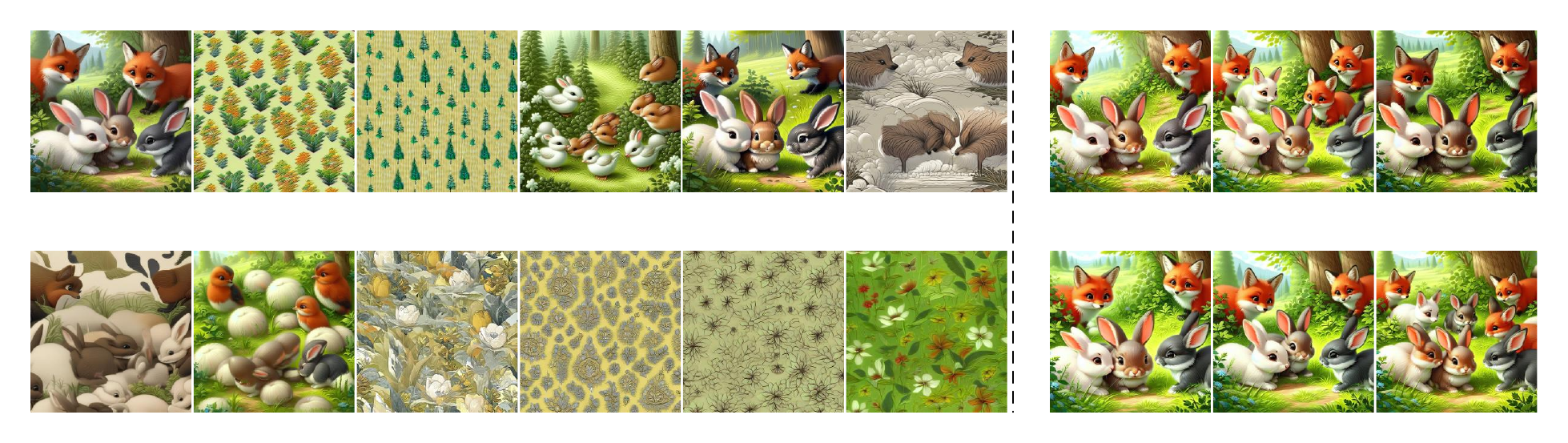}
		\put(71.5,13){\footnotesize\parbox{4cm}{{{\textit{“$\varnothing$”}}}}}
		\put(65.7,15.8){\footnotesize\rotatebox{90}{\parbox{4cm}{{\textcolor{red}{$\downarrow$}{\{$B_3 B_4 B_6$\}}}}}}
		\put(77.9,13){\footnotesize\parbox{4cm}{{{\textit{“...cartoon\ style”}}}}}
		\put(65.7,1.5){\footnotesize\rotatebox{90}{\parbox{4cm}{{\textcolor{red}{$\downarrow$}{\{$B_3 B_4 B_6$\}}}}}}
		\put(90.5,13){\footnotesize\parbox{1.5cm}{{{\textit{“...dark horror style”}}}}}
		\put(68.2,-1.8){\footnotesize\parbox{1.5cm}{{{\textit{“...in a natural field”}}}}}
		\put(78.3,-1.8){\footnotesize\parbox{4cm}{{{\textit{“...in the forest”}}}}}
		\put(89.5,-1.8){\footnotesize\parbox{4cm}{{{\textit{“...voxels style”}}}}}
		\put(2.2,12.3){\footnotesize\parbox{4cm}{{\textcolor{red}{$\downarrow$}{\{$B_6$\}}}}}
		\put(13.7,12.3){\footnotesize\parbox{4cm}{{\textcolor{red}{$\downarrow$}{\{$B_7$\}}}}}
		\put(24.5,12.3){\footnotesize\parbox{4cm}{{\textcolor{red}{$\downarrow$}{\{$B_8$\}}}}}
		\put(35,12.3){\footnotesize\parbox{4cm}{{\textcolor{red}{$\downarrow$}{\{$B_9$\}}}}}
		\put(46,12.3){\footnotesize\parbox{4cm}{{\textcolor{red}{$\downarrow$}{\{$B_{10}$\}}}}}
		\put(56.8,12.3){\footnotesize\parbox{4cm}{{\textcolor{red}{$\downarrow$}{\{$B_{11}$\}}}}}
		\put(2.2,27){\footnotesize\parbox{4cm}{{{Reference}}}}
		\put(13.7,27){\footnotesize\parbox{4cm}{{\textcolor{red}{$\downarrow$}{\{$B_1$\}}}}}
		\put(24.5,27){\footnotesize\parbox{4cm}{{\textcolor{red}{$\downarrow$}{\{$B_2$\}}}}}
		\put(35,27){\footnotesize\parbox{4cm}{{\textcolor{red}{$\downarrow$}{\{$B_3$\}}}}}
		\put(46,27){\footnotesize\parbox{4cm}{{\textcolor{red}{$\downarrow$}{\{$B_{4}$\}}}}}
		\put(57,27){\footnotesize\parbox{4cm}{{\textcolor{red}{$\downarrow$}{\{$B_{5}$\}}}}}
		\put(0,-1.8){\footnotesize\parbox{4cm}{{{(a)}}}}
		\put(65.7,-1.8){\footnotesize\parbox{4cm}{{{(b)}}}}
	\end{overpic}
	\vspace{-3pt}
	\caption{{Effects of different insertion positions}. Reference denotes the input image. \textcolor{red}{$\downarrow$}{{$B_i$}} indicates the insertion of the Quantity Attention (QTTN) module into block $B_i$ of the U-Net, while $\varnothing$ signifies that the user text prompt is set to null-text. (a) The first and second rows show the results obtained by inserting the QTTN module into different blocks of the U-Net during training and applying null-text during inference. When the input image is inserted into $B_4$, it effectively utilizes the quantity perception from the QTTN module. (b) The first and second rows demonstrate the image editing effects achieved by training with the QTTN module inserted into three blocks ($B_3$, $B_4$, $B_6$) simultaneously and applying varying user text prompts during inference.}
	\label{insert_position}
\end{figure*}
\begin{figure}[t]
	\centering
	\begin{overpic}[width=0.48\textwidth, trim=20 20 20 15, clip]{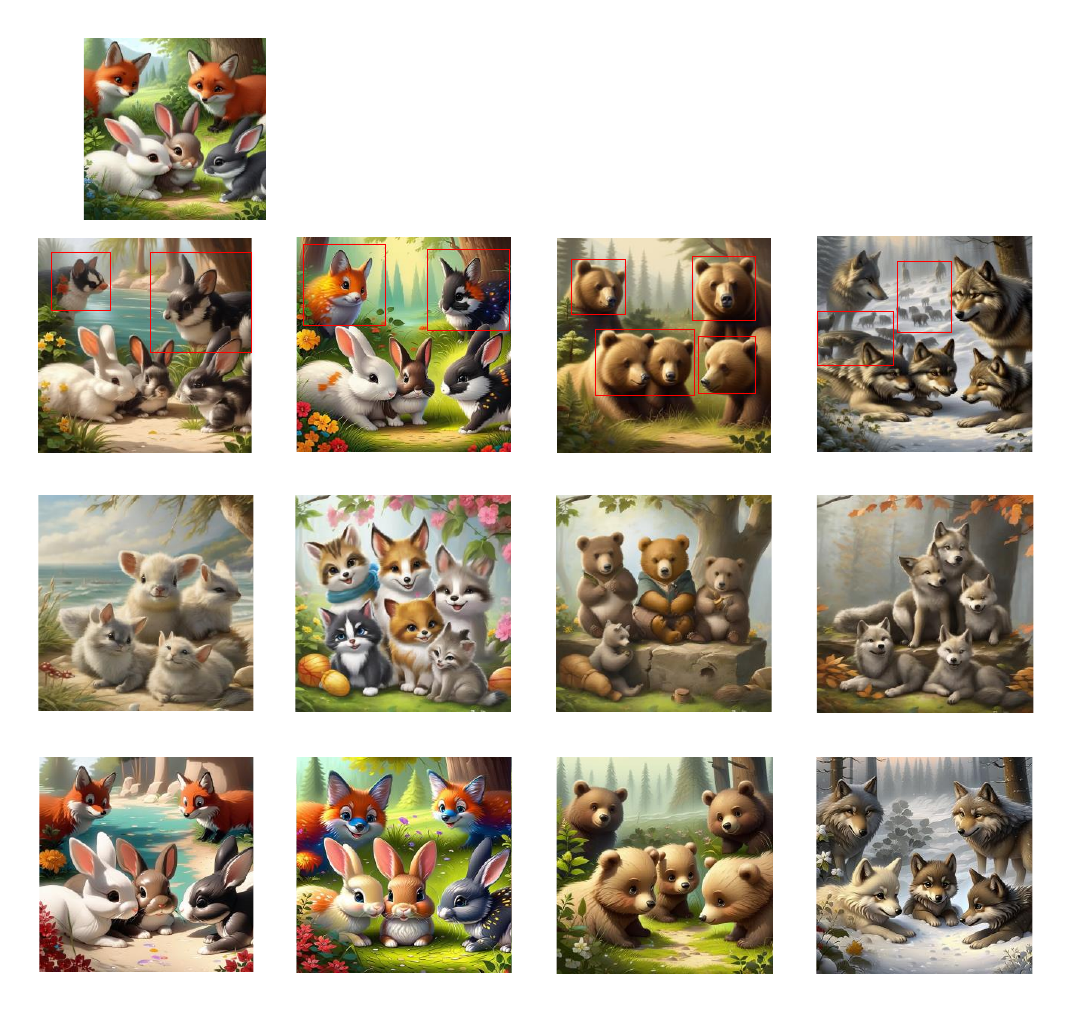}
		\put(28,84){\parbox{5cm}{\centering\textcolor{lightblue}{\textit{three rabbits and two foxes}}}}
		\put(-1,-5){\footnotesize\parbox{2cm}{\centering{{\textit{“...by the sea”}}}}}
		\put(24.5,-5){\footnotesize\parbox{2cm}{\centering{{\textit{“...colorful style”}}}}}
		\put(51.2,-5){\footnotesize\parbox{2cm}{\centering{{\textit{“\textcolor{red}{$\rightarrow$}bears”}}}}}
		\put(77,-5){\footnotesize\parbox{2cm}{\centering{{\textit{“\textcolor{red}{$\rightarrow$}wolves, in the snow”}}}}}
		\put(20.2,22.5){\scriptsize\parbox{5cm}{\centering{{{MoEdit(Ours)}}}}}
		\put(20.2,49.2){\scriptsize\parbox{5cm}{\centering{{{w/o QTTN module}}}}}
		\put(20,75.5){\scriptsize\parbox{5cm}{\centering{{{w/o FeCom module}}}}}
		\put(-10.5,84){\small\parbox{2cm}{\centering{{\rotatebox{90}{Reference}}}}}
	\end{overpic}
	\vspace{-1pt}
	\caption{{Ablation study.} The first row demonstrates that the exclusion of the Feature Compensation (FeCom) module results in significant instability in the visual representations of object attributes and image details. The second row demonstrates that the absence of the Quantity Attention (QTTN) module causes MoEdit to completely lose its capability of consistent quantity perception, leading to substantial disruptions in image structure and semantics. The final row showcases the full capability of MoEdit, which effectively completes editing tasks with high quality, preserving quantity consistency and extracting object attributes.}
	\label{ablation study}
	\vspace{-5pt}
\end{figure}
\subsection{Implementation}
The experiments are conducted using PyTorch on NVIDIA A6000 GPUs. This research utilized SDXL \cite{podell2023sdxl} as the baseline model and employed a CLIP model \cite{Radford2021LearningTV} pre-trained on the Laion-2B \cite{schuhmann2022laion} as the image encoder. During training, all pre-trained weights remain frozen, while only the FeCom and QTTN module weights are fine-tuned. The scaling factors $\lambda$ and $\beta$ are both set to 1.0, with a learning rate of 0.00025. The Adam optimizer is used to perform optimization over 12,000 steps. Notably, MoEdit also supports single-image training and can perform editing on this image.
Regarding the dataset, it comprises over 1,200 multi-object images (3 to 14 objects and 3 to 5 $c_q$ per image). For evaluation, 30 test images are selected with 50 to 150 editing instructions per image, generating approximately 3,000 result images for quantitative evaluation. Selected results are analyzed in subsequent sections. The dataset will be expanded and made publicly available for future research.
\subsection{Quantitative Comparisons}
Seven algorithms are selected for comparison: SSR-Encoder \cite{zhang2024ssr}, $\lambda$-Eclipse \cite{patel2024lambda}, IP-Adapter \cite{ye2023ip}, Blip-diffusion \cite{li2024blip}, MS-diffusion \cite{wang2024ms}, Emu2 \cite{sun2024generative}, and TurboEdit \cite{wu2024turboedit}, as summarized in Table~\ref{tab:comparison attribute}. Six objective metrics are employed: HyperIQA \cite{su2020blindly}, AesBench \cite{huang2024aesbench}, Q-Align \cite{wu2023q}, NIQE \cite{mittal2012making}, CLIP Score \cite{hessel2021clipscore}, and LPIPS \cite{zhang2018unreasonable}. These metrics evaluate various aspects: Perceptual Quality: NIQE, HyperIQA, and LPIPS focus on naturalness and visual fidelity.
Semantic Consistency: CLIP Score quantifies alignment between generated images and text prompts.
Visual Composition and Aesthetics: Q-Align and AesBench assess image composition and aesthetic quality.
Additionally, two subjective metrics are adopted: Numerical Accuracy, measuring the percentage of successfully preserving quantity in the results, and MOS, reflecting the proportion of participants favoring a specific model. To collect subjective evaluations, 20 users are invited in image evaluation.

A comprehensive evaluation is provided in Table~\ref{tab:objective metrics}. While TurboEdit achieves the highest LPIPS score, indicating strong perceptual similarity to original images, MoEdit outperforms all methods on the remaining metrics. On average, MoEdit demonstrates a 3.83\% improvement across all metrics, with the most significant gain observed in aesthetic performance under Q-Align (5.82\%) and the smallest in NIQE (0.79\%). In subjective evaluations, MoEdit slightly lags behind Emu2 in Numerical Accuracy due to the clustering strategy of Emu2, which enhances numerical stability in visualization but restricts editability. Overall, MoEdit excels in meeting user expectations and aesthetic preferences by extracting object attributes and ensuring consistent quantity perception during editing.
\vspace{0pt}
\subsection{Qualitative Comparisons} 
\vspace{0pt}
Fig.~\ref{comparisons} illustrates qualitative comparisons between MoEdit and seven competing methods on multi-object image editing tasks. Significant limitations are observed in SSR-Encoder, $\lambda$-Eclipse, and Blip-diffusion, including limited editability, instability in object attributes and quantity perception, image distortion, and inconsistency in structural and semantic alignment. While IP-Adapter shows slight advantages in clarity and realism due to its robust baseline model, it falls short of MoEdit.
MS-diffusion demonstrates reasonable editability and quantity preservation when dealing with a small number of objects but struggles with larger numbers, often misinterpreting numerical constraints and object attributes. Emu2 maintains quantity and structural consistency well but frequently suffers from image distortion and over-clustering, limiting diversity and editability. TurboEdit achieves relatively balanced performance but trails MoEdit in critical aspects such as quantity preservation, editability, and handling complex style modifications.
In contrast, MoEdit consistently delivers superior results, balancing semantic accuracy, stylistic consistency, and aesthetic quality. Its ability to extract object attributes and preserve quantity consistency while fulfilling user expectations sets it apart from other methods. For more qualitative demonstrations, please refer to {Supplementary Material}.
\begin{table}[]
	\renewcommand{\arraystretch}{1.2}
	\scriptsize
	\resizebox{\columnwidth}{!}{%
		\begin{tabular}{cccccc}
			\hline
			\multirow{3}{*}{\textbf{QTTN}} & \multirow{3}{*}{\textbf{FeCom}} & \multirow{3}{*}{\textbf{NIQE $\downarrow$}}  & \multirow{3}{*}{\textbf{HyperIQA $\uparrow$}}  & \multicolumn{2}{c}{\multirow{2}{*}{\textbf{CLIP Score}}} \\
			&                                 &                                 &                                     & \multicolumn{2}{c}{}                                     \\ \cline{5-6} 
			&                                 &                                 &                                     & \textbf{Whole $\uparrow$}             & \textbf{Edit $\uparrow$}               \\ \hline
			\ding{55}                              & \ding{51}                               & 2.9988                               & 71.82                                   & 0.2831                          & 0.2687                           \\
			\ding{51}                              & \ding{55}                               & 2.6890                               & 76.17                                   & 0.3052                          & 0.2752                           \\
			\ding{51}                              & \ding{51}                               & \textbf{2.6501}                               & \textbf{77.87}                                   & \textbf{0.3274}                          & \textbf{0.2790}                           \\ \hline \hline
			\multirow{2}{*}{\textbf{QTTN}} & \multirow{2}{*}{\textbf{FeCom}} & \multirow{2}{*}{\textbf{LPIPS $\downarrow$}} & \multirow{2}{*}{\textbf{Numerical $\uparrow$}} & \multicolumn{2}{c}{\textbf{Q-Align}}                     \\ \cline{5-6} 
			&                                 &                                 &                                     & \textbf{Quality $\uparrow$}           & \textbf{Aesthetic $\uparrow$}          \\ \hline
			\ding{55}                              & \ding{51}                               & 0.3116                               & 25.67                                   & 4.8945                          & 4.4602                           \\
			\ding{51}                              & \ding{55}                               & 0.2777                               & 82.31                                   & 4.9023                          & 4.6562                           \\
			\ding{51}                              & \ding{51}                               & \textbf{0.2731}                               & \textbf{86.79}                                   & \textbf{4.9219}                          & \textbf{4.8047}                           \\ \hline
		\end{tabular}%
	}
	\vspace{-3pt}
	\caption{{Ablation study.} Removing the FeCom module significantly reduces the whole metric in CLIP Score and aesthetics metric in Q-Align, due to the text-image mismatches and aliasing of object attributes, respectively, leading to failures in extracting object attributes. In contrast, removing the QTTN module causes declines across all metrics, including quantity preservation, structural consistency, editability, and aesthetics. These results demonstrate that the two modules are complementary to achieve the optimal performance of MoEdit.}
	\label{tab:ablation study}
\end{table}
\begin{figure}[t]
	\centering
	\begin{overpic}[width=0.48\textwidth, trim=20 20 20 15, clip]{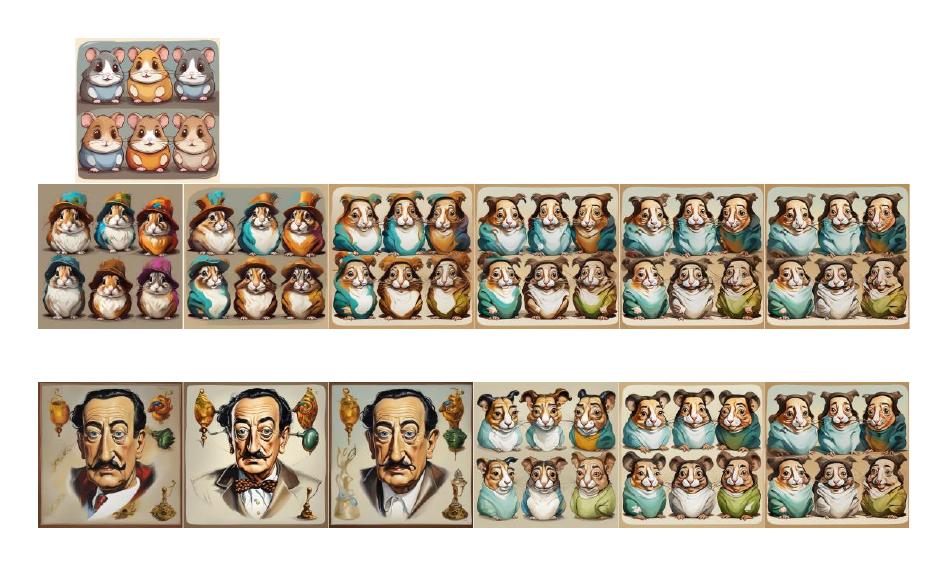}
		\put(26,47){\small\parbox{5cm}{\centering\textcolor{lightblue}{\textit{six hamsters}}\textit{\ vibrant portrait painting of Salvador Dali}}}
		\put(-10.5,47){\small\parbox{2cm}{\centering{{\rotatebox{90}{Reference}}}}}
		\put(-4,-3.5){\scriptsize\parbox{2cm}{\centering{{{($\beta$\ =\ 0, $\lambda$\ =\ 1)}}}}}
		\put(13,-3.5){\scriptsize\parbox{2cm}{\centering{{{(0.2, 1)}}}}}
		\put(29.5,-3.5){\scriptsize\parbox{2cm}{\centering{{{(0.4, 1)}}}}}
		\put(46.5,-3.5){\scriptsize\parbox{2cm}{\centering{{{(0.6, 1)}}}}}
		\put(63,-3.5){\scriptsize\parbox{2cm}{\centering{{{(0.8, 1)}}}}}
		\put(79.5,-3.5){\scriptsize\parbox{2cm}{\centering{{{(1, 1)}}}}}
		\put(-4,18.5){\scriptsize\parbox{2cm}{\centering{{{($\beta$\ =\ 1, $\lambda$\ =\ 0)}}}}}
		\put(13,18.5){\scriptsize\parbox{2cm}{\centering{{{(1, 0.2)}}}}}
		\put(29.5,18.5){\scriptsize\parbox{2cm}{\centering{{{(1, 0.4)}}}}}
		\put(46.5,18.5){\scriptsize\parbox{2cm}{\centering{{{(1, 0.6)}}}}}
		\put(63,18.5){\scriptsize\parbox{2cm}{\centering{{{(1, 0.8)}}}}}
		\put(79.5,18.5){\scriptsize\parbox{2cm}{\centering{{{(1, 1)}}}}}
		%\put(51.2,-2){\tiny\parbox{2cm}{\centering{{\textit{“\textcolor{red}{$\rightarrow$}bears”}}}}}
	\end{overpic}
	\vspace{-7pt}
	\caption{{Scale variation}. The first and second rows illustrate the variations in editing results when Reference and \textit{“vibrant portrait painting of Salvador Dalí”} are used as inputs, under different $\lambda$ and $\beta$ values. The numerical values below each image represent the corresponding $\lambda$ and $\beta$ settings. As $\lambda$ decreases, the objects progressively lose the attributes of the \textit{“vibrant portrait painting of Salvador Dalí”}. Similarly, as beta decreases, MoEdit gradually loses its sensitivity to quantitative attributes.}
	\label{scale_adjust}
\end{figure}
\subsection{Ablation Study}
\label{sec:ablation study}
\noindent\textbf{Insertion point.}
%\paragraph{Insertion point.}
Fig.~\ref{insert_position}~(a) illustrates the effectiveness of integrating quantity perception into U-Net when the QTTN module is inserted into a single block. Specifically, when the QTTN module is incorporated into $B_4$, U-Net effectively leverages quantity perception to enhance performance. In contrast, Fig.~\ref{insert_position}~(b) demonstrates the impact of injecting quantity perception into multiple blocks simultaneously. This approach disrupts the guidance provided by text prompts, significantly compromising editability and stability. Consequently, integrating the QTTN module exclusively into $B_4$ achieves an optimal trade-off between utilizing quantity perception and maintaining editability.\\
\noindent\textbf{Module Effectiveness.}
%\paragraph{Module Effectiveness.}
Excluding the FeCom module (“w/o FeCom module”) from the pipeline leads to significant performance degradation, as shown in the first row of Fig.~\ref{ablation study}. 
The first and second columns reveal a severe aliasing of object attributes, while the third column highlights the failure to extract the inherent juvenile characteristics of the entities in the input image. Together, these observations suggest an inability to effectively extract object attributes. Furthermore, the fourth column demonstrates a notable degradation of details in the output image.
Similarly, removing the QTTN module (“w/o QTTN module”) produces results displayed in the second row of Fig.~\ref{ablation study}. The absence of the QTTN module impairs quantity perception, resulting in inconsistencies in object counts, image structures, and object semantics.

Table~\ref{tab:ablation study} quantifies these effects. Removing the FeCom module significantly reduces the whole and aesthetic metric in CLIP Score and Q-Align, respectively, primarily due to mismatches between object attributes and textual prompts, object attributes aliasing, and loss of image detail. In contrast, removing the QTTN results in notable declines across all metrics, including quantity perception, image quality, structural consistency, editability, and aesthetics. These findings highlight the complementary roles of the FeCom and QTTN in optimizing the performance of MoEdit.\\
\vspace{0pt}
%\paragraph{Adjustable Scale Variation.}
\noindent\textbf{Adjustable Scale Variation.}
Fig.~\ref{scale_adjust} depicts how variations in $\lambda$ and $\beta$ during inference affect editing outcomes. Gradually reducing $\lambda$ diminishes the \textit{“vibrant portrait painting of Salvador Dalí”} attribute in the six hamsters, which vanishes entirely when $\lambda$ reaches 0. Similarly, decreasing $\beta$ weakens the capability of quantity perception of QTTN module, leading to increased structural and semantic disorder in the generated images. When $\beta$ is reduced to 0, the generated images lose structural and semantic coherence entirely.
\vspace{-2pt}
\section{Conclusion}
This paper presents MoEdit, an novel auxiliary-free framework for multi-object image editing that addresses critical challenges in quantity perception and object attribute extraction. By integrating the FeCom and QTTN modules, MoEdit achieves SOTA performance in editing quality, object attributes extraction, and consistent quantity preservation. The FeCom module effectively reduces aliasing of object attributes, while the QTTN module ensures consistent quantity perception without reliance on auxiliary tools. Experimental results demonstrate the superiority of MoEdit over existing methods in terms of visual quality, semantic consistency, and editability. Future work can enhance the model robustness in complex interaction scenarios.
\vspace{-2pt}
\section*{Acknowledgment}
The work was supported in part by the National Natural Science Foundation of China under Grant 62301310 and 62176169, Sichuan Science and Technology Program under Grant 2024NSFSC1426, Science and Technology Development Fund of Macao Grant (0004/2024/E1B1) and Macao Polytechnic University Grant (RP/FCA-10/2023).

% \section{Acknowledgement}
% This work was supported in part by NSFC (No.62225112, No.61831015), the Fundamental Research Funds for the Central Universities, National Key R\&D Program of China 2021YFE0206700, and Shanghai Municipal Science and Technology Major Project (2021SHZDZX0102).

{
    \small
    \bibliographystyle{ieeenat_fullname}
    \bibliography{main}
}

\clearpage
\setcounter{page}{1}
\maketitlesupplementary

\appendix

\section{Details of Network Architecture}
\label{sec:Details of Network Architecture}
This section provides a detailed explanation of the design of two key modules in MoEdit: the Feature Compensation (FeCom) module and the Quantity Attention (QTTN) module. It covers their structural composition, dimensional transformations, inputs, and outputs. The FeCom module is designed to minimize interlacing during the extraction of object attributes, while the QTTN module ensures consistent quantity preservation.
\subsection{Feature Compensation Module}
\label{sec:Feature Compensation Module}
\subsubsection{Overall}
\label{sec:Overall-FeCom}
\begin{figure}[htbp]
	\centering
	\begin{overpic}[width=0.48\textwidth, trim=18 10 18 15, clip]{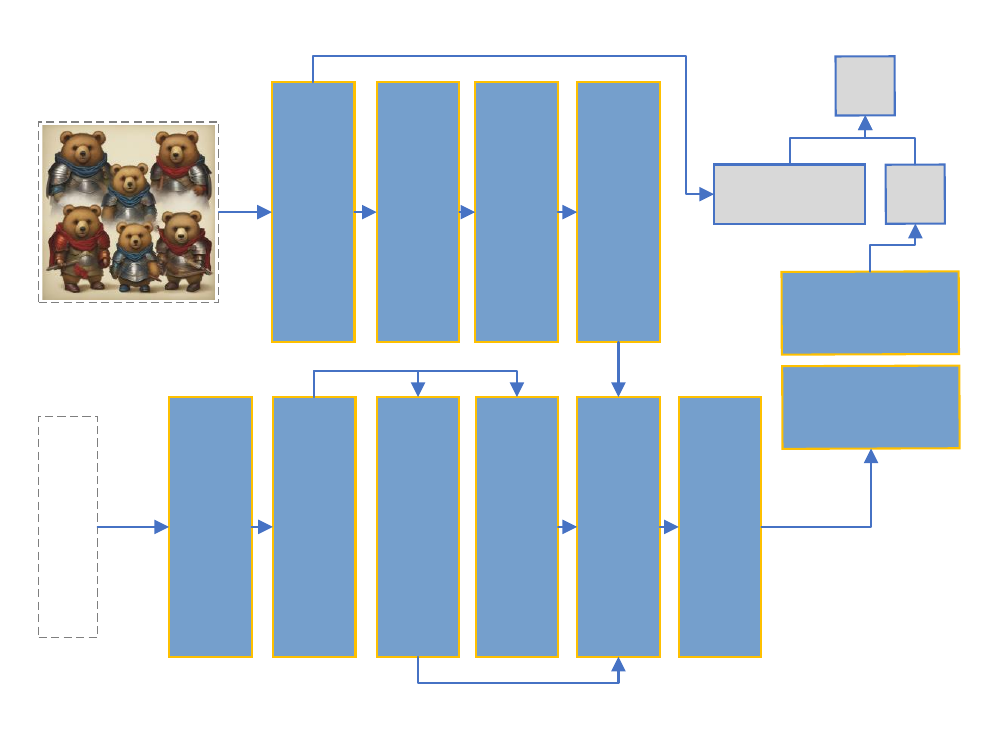}
		\put(1.8,9.5){\rotatebox{90}{\parbox{4cm}{\textcolor{lightblue}{six doll bears}}}}
		\put(15.5,10.7){\small\rotatebox{90}{\parbox{4cm}{\textcolor{white}{Text Encoder}}}}
		\put(19.8,12.5){\tiny\rotatebox{90}{\parbox{4cm}{\textcolor{white}{$\rightarrow$ $(77, 2048)$}}}}
		\put(27,14){\small\rotatebox{90}{\parbox{4cm}{\textcolor{white}{FC Layer}}}}
		\put(31,12.5){\tiny\rotatebox{90}{\parbox{4cm}{\textcolor{white}{$\rightarrow$ $(77, 2048)$}}}}
		\put(38,11){\small\rotatebox{90}{\parbox{4cm}{\textcolor{white}{K Projection}}}}
		\put(42.5,11.8){\tiny\rotatebox{90}{\parbox{4cm}{\textcolor{white}{$\rightarrow$ $(10,77,64)$}}}}
		\put(48.7,11){\small\rotatebox{90}{\parbox{4cm}{\textcolor{white}{V Projection}}}}
		\put(53,11.8){\tiny\rotatebox{90}{\parbox{4cm}{\textcolor{white}{$\rightarrow$ $(10,77,32)$}}}}
		\put(59.8,9){\small\rotatebox{90}{\parbox{4cm}{\textcolor{white}{Cross Attention}}}}
		\put(64,12.2){\tiny\rotatebox{90}{\parbox{4cm}{\textcolor{white}{$\rightarrow$ $(10,4,32)$}}}}
		\put(70.8,15.5){\small\rotatebox{90}{\parbox{4cm}{\textcolor{white}{Resize}}}}
		\put(75.2,15){\tiny\rotatebox{90}{\parbox{4cm}{\textcolor{white}{$\rightarrow$ $(1280)$}}}}
		
		\put(27,43){\small\rotatebox{90}{\parbox{4cm}{\textcolor{white}{Image Encoder}}}}
		\put(31,48.5){\tiny\rotatebox{90}{\parbox{4cm}{\textcolor{white}{$\rightarrow$ $(1280)$}}}}
		\put(38,48){\small\rotatebox{90}{\parbox{4cm}{\textcolor{white}{FC Layer}}}}
		\put(42,48.5){\tiny\rotatebox{90}{\parbox{4cm}{\textcolor{white}{$\rightarrow$ $(2560)$}}}}
		\put(49,50){\small\rotatebox{90}{\parbox{4cm}{\textcolor{white}{Resize}}}}
		\put(53,48){\tiny\rotatebox{90}{\parbox{4cm}{\textcolor{white}{$\rightarrow$ $(4,640)$}}}}
		\put(59.8,45){\small\rotatebox{90}{\parbox{4cm}{\textcolor{white}{Q Projection}}}}
		\put(64,46){\tiny\rotatebox{90}{\parbox{4cm}{\textcolor{white}{$\rightarrow$ $(10,4,64)$}}}}
		
		\put(83,34.5){\footnotesize{\parbox{4cm}{\textcolor{white}{LayerNorm}}}}
		\put(84.5,31){\tiny{\parbox{4cm}{\textcolor{white}{$\rightarrow$ $(1280)$}}}}
		\put(83.8,44.5){\footnotesize{\parbox{4cm}{\textcolor{white}{FC Layer}}}}
		\put(84.5,41){\tiny{\parbox{4cm}{\textcolor{white}{$\rightarrow$ $(1280)$}}}}
		\put(93.7,55.5){\footnotesize{\parbox{4cm}{\textcolor{black}{$I_c$}}}}
		\put(74.8,55.5){\scriptsize{\parbox{4cm}{\textcolor{black}{$CLIP(I)$}}}}
		\put(88.2,67.2){\footnotesize{\parbox{4cm}{\textcolor{black}{$I_g$}}}}
	\end{overpic}
	\caption{The detailed structure diagram of FeCom module. The image encoder is derived from CLIP \cite{radford2021learning}, and the text encoder is sourced from SDXL \cite{podell2023sdxl}. The module takes two inputs—images and text prompts—and generates a single output: enhanced features $I_g$. In this process, $CLIP(I)$ denotes the inferior image features extracted by the image encoder of CLIP, while $I_c$ represents the compensation features used to enhance it. Each step is explicitly defined, with “$\rightarrow$ (...)” denoting the dimensionality of the corresponding output.}
	\label{fig:feature attention}
\end{figure}

This module processes an input image $I$ and a text prompt $c_q$, which specifies objects and their quantities, to generate an enhanced feature $I_g$, characterized by distinct and separable object attributes. The module is composed of three main components: an image encoder (from CLIP \cite{radford2021learning}), a text encoder (from SDXL \cite{podell2023sdxl}), and a feature attention mechanism. The feature attention mechanism aligns textual and visual representations of object and quantity information, enabling effective extraction of object attributes. Fig.~\ref{fig:feature attention} provides a detailed overview of the FeCom module, with each step annotated to indicate its function. The notation “$\rightarrow$ (...)” represents the output dimensions at each stage.
\begin{figure*}[htbp]
	\centering
	\begin{overpic}[width=1\textwidth, trim=18 10 18 15, clip]{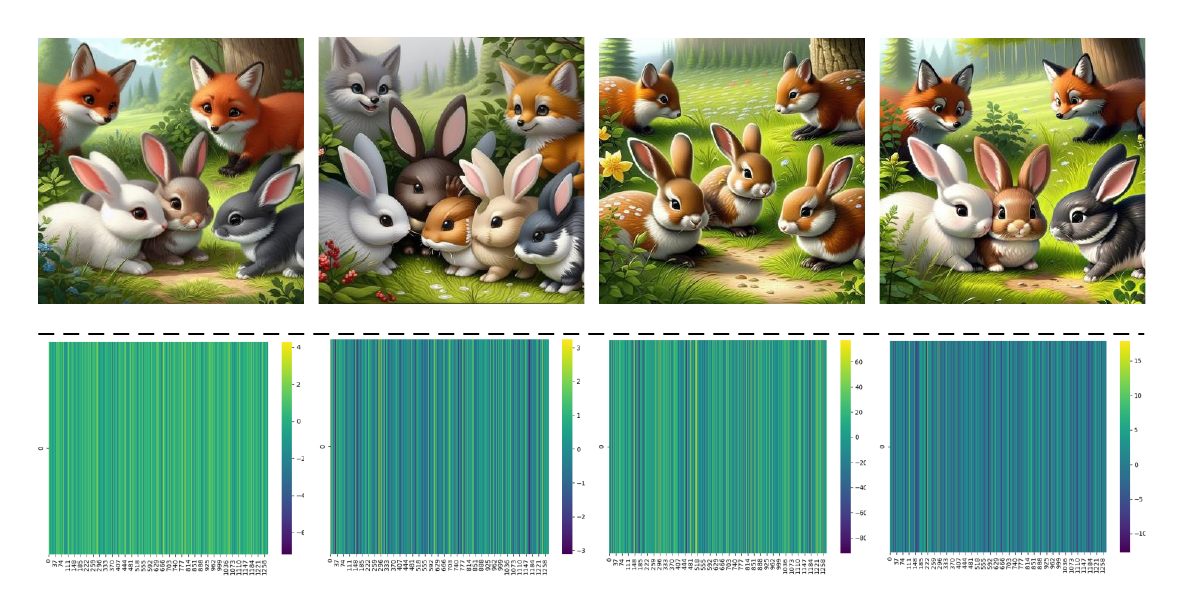}
		\put(0,24){{\footnotesize\parbox{4cm}{{(a)}}}}
		\put(8.5,50.5){{\parbox{4cm}{{Reference}}}}
		\put(59.5,50.5){{\parbox{4cm}{{Output}}}}
		\put(32,24){{\footnotesize\parbox{4cm}{m/o LayerNorm}}}
		\put(57.5,24){{\footnotesize\parbox{4cm}{w/o LayerNorm}}}
		\put(83,24){{\footnotesize\parbox{4cm}{MoEdit (Ours)}}}
		\put(3.5,0){{\footnotesize\parbox{4cm}{Original Image Features}}}
		\put(0,0){{\footnotesize\parbox{4cm}{(b)}}}
		\put(27,0){{\footnotesize\parbox{4cm}{Compensation Features (m/o)}}}
		\put(52,0){{\footnotesize\parbox{4cm}{Compensation Features (w/o)}}}
		\put(77,0){{\footnotesize\parbox{4cm}{Compensation Features (Ours)}}}
	\end{overpic}
	\caption{The importance of the LayerNorm layer. Please refer to Sec.~\ref{sec:Discussion of LayerNorm} for a comprehensive evaluation and detailed discussions regarding these configurations.}
	\label{fig:layernorm}
\end{figure*}
\subsubsection{Discussion of LayerNorm}
\label{sec:Discussion of LayerNorm}
The FeCom module incorporates a LayerNorm layer, which is critical for balancing original image features, aligning textual and visual information, and enabling the QTTN module to effectively consider each object both individually and part of the whole image. Fig.~\ref{fig:layernorm} underscores its significance, with Reference representing the input image.

In Fig.~\ref{fig:layernorm} (a), “m/o LayerNorm” denotes the results obtained when the LayerNorm layer from Fig.~\ref{fig:feature attention} is repositioned to the final stage, after the last fully connected (FC) layer, during training. This adjustment leads to a marked decline in both quantity consistent perception and object attributes extraction. Conversely, “w/o LayerNorm” refers to the scenario where the LayerNorm layer is entirely removed during training. While this configuration preserves quantity consistency, it introduces aliasing of object attributes.

In Fig.~\ref{fig:layernorm} (b), to elucidate these degradations, we analyze the original image features extracted by the image encoder of CLIP and the compensation features generated by the FeCom module. “Compensation Features (m/o)” refers to compensation features obtained under the “m/o LayerNorm” condition, with a value range of $[3, -3]$, while the original image features span $[4, -6]$. This narrow range suggests insufficient feature intensity, which fails to adequately mitigate the in-between interlacing in inferior original image features, thereby compromising quantity consistent perception and object attributes extraction. Conversely, “Compensation Features (w/o)” are produced under the “w/o LayerNorm” condition, with a value range of $[60, -80]$. This excessively broad range indicates over-suppression of original image features, resulting in substantial information loss, particularly in object attributes. In comparison, the compensation features constructed by the FeCom module in MoEdit (denoted as “Compensation Features (Ours)”) fall within a balanced range of $[15, -10]$. This range effectively preserves the original image information while minimizing in-between interlacing. 

The quantitative comparisons are shown in Table~\ref{tab:function of layernorm and extraction}. Notably, the “m/o LayerNorm” have a significant impact on the overall performance of MoEdit, primarily due to their limited compensatory capacity. In contrast, under the “w/o LayerNorm” condition, despite the excessive strength of the compensation features, image aesthetics, quality, and numerical accuracy remain largely unaffected. This stability can be attributed to the fact that these compensation features are derived from textual prompts containing object and quantity information. However, the loss of object attributes leads to a noticeable decline in textual-visual alignment and the similarity between the original and resulting images.

As demonstrated in Fig.~\ref{fig:layernorm} (a), MoEdit achieves superior performance visually in both preserving quantity consistency and extracting object attributes.
\subsection{Quantity Attention Module}
\label{sec:Quantity Attention Module}
\begin{table}[]
	\centering
	\renewcommand{\arraystretch}{1.2}
	\resizebox{\columnwidth}{!}{%
		\begin{tabular}{ccccc}
			\hline
			\multirow{2}{*}{\textbf{Method}} & \multirow{2}{*}{\textbf{NIQE} $\downarrow$}  & \multirow{2}{*}{\textbf{HyperIQA} $\uparrow$}  & \multicolumn{2}{c}{\textbf{CLIP Score}} \\ \cline{4-5} 
			&                        &                            & \textbf{Whole} $\uparrow$        & \textbf{Edit} $\uparrow$          \\ \hline
			m/o LayerNorm           & 2.8726                       & 73.28                           & 0.3005              &  0.2689              \\
			w/o LayerNorm           & 2.7081                       & 75.99                           & 0.3078              &  0.2778              \\
			w/o Extraction          & 2.7232                       & 75.47                           & 0.3105              &  0.2782              \\
			MoEdit (Ours)           & \textbf{2.6501}                               & \textbf{77.87}                                   & \textbf{0.3274}                          & \textbf{0.2790}                \\ \hline \hline
			\multirow{2}{*}{\textbf{Method}} & \multirow{2}{*}{\textbf{LPIPS} $\downarrow$} & \multirow{2}{*}{\textbf{Numerical} $\uparrow$} & \multicolumn{2}{c}{\textbf{Q-Align}}    \\ \cline{4-5} 
			&                        &                            & \textbf{Quality} $\uparrow$      & \textbf{Aesthetic} $\uparrow$     \\ \hline
			m/o LayerNorm           & 0.3101                       & 32.65                           & 4.7511              & 4.4371               \\
			w/o LayerNorm           & 0.2943                       & 84.77                           & 4.8933              & 4.7552               \\
			w/o Extraction          & 0.2975                   &  78.73                          &  4.8344             &  4.6211              \\
			MoEdit (Ours)           & \textbf{0.2731}                               & \textbf{86.79}                                   & \textbf{4.9219}                          & \textbf{4.8047}               \\ \hline
		\end{tabular}%
	}
	\caption{Quantitative comparisons. “m/o LayerNorm” denotes the results obtained when the LayerNorm layer from Fig.~\ref{fig:feature attention} is repositioned to the final stage, after the last FC layer, during training. “w/o LayerNorm” refers to the scenario where the LayerNorm layer is entirely removed during training. “w/o Extraction” illustrates the absence of Extraction module in Fig.~\ref{fig:quantity attention}. For more discussion, please refer to Secs.~\ref{sec:Discussion of LayerNorm} and ~\ref{sec:Discussion of Extraction}.}
	\label{tab:function of layernorm and extraction}
\end{table}
\subsubsection{Overall}
\label{sec:Overall-QTTN}
\begin{figure}[htbp]
	\centering
	\begin{overpic}[width=0.48\textwidth, trim=18 10 18 15, clip]{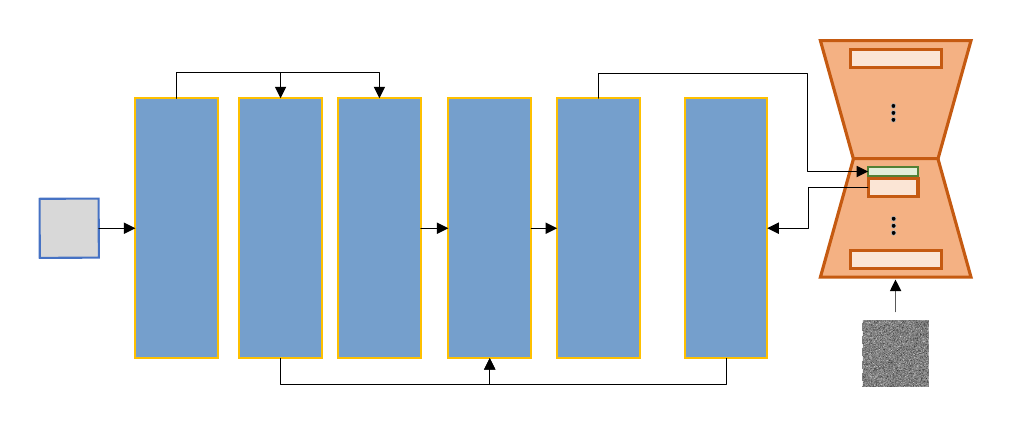}
		\put(2,19){\small{\parbox{4cm}{\textcolor{black}{$I_g$}}}}
		\put(-0.5,14.5){\tiny{\parbox{4cm}{\textcolor{black}{$(1280)$}}}}
		\put(11.8,12){\small\rotatebox{90}{\parbox{4cm}{\textcolor{white}{Extraction}}}}
		\put(16,13.5){\tiny\rotatebox{90}{\parbox{4cm}{\textcolor{white}{$\rightarrow$ $(1280)$}}}}
		\put(22.8,10.5){\small\rotatebox{90}{\parbox{4cm}{\textcolor{white}{K Projection}}}}
		\put(27,11.5){\tiny\rotatebox{90}{\parbox{4cm}{\textcolor{white}{$\rightarrow$ $(20, 4, 64)$}}}}
		\put(33.3,10.5){\small\rotatebox{90}{\parbox{4cm}{\textcolor{white}{V Projection}}}}
		\put(37.5,11.5){\tiny\rotatebox{90}{\parbox{4cm}{\textcolor{white}{$\rightarrow$ $(20, 4, 64)$}}}}
		\put(45,8.5){\small\rotatebox{90}{\parbox{4cm}{\textcolor{white}{Cross Attention}}}}
		\put(49.5,10){\tiny\rotatebox{90}{\parbox{4cm}{\textcolor{white}{$\rightarrow$ $(20, 1024, 64)$}}}}
		\put(56.5,15){\small\rotatebox{90}{\parbox{4cm}{\textcolor{white}{Resize}}}}
		\put(60.5,10.5){\tiny\rotatebox{90}{\parbox{4cm}{\textcolor{white}{$\rightarrow$ $(1024, 1280)$}}}}
		\put(70.2,10.5){\small\rotatebox{90}{\parbox{4cm}{\textcolor{white}{Q Projection}}}}
		\put(74.5,10.5){\tiny\rotatebox{90}{\parbox{4cm}{\textcolor{white}{$\rightarrow$ $(1024, 1280)$}}}}
		\put(67,38.5){\small{\parbox{4cm}{\textcolor{black}{Inject}}}}
		\put(79,16){\small{\parbox{4cm}{\textcolor{black}{$z_t^4$}}}}
	\end{overpic}
	\caption{The detailed structure diagram of QTTN module. Each step is annotated with its specific purpose, with “$\rightarrow$ (...)” denoting the dimensionality of the corresponding output. This module receives two inputs: $I_g$, originating from the FeCom module, and input noise $z_t^4$, from the fourth block $B_4$ of the U-Net. Its primary purpose is to generate a quantity consistent perception that can be effectively interpreted by the U-Net. The core of the Extraction module $E_t$ is a FC layer, which is designed to consider each object both individually and part of the whole $I_g$. The functionality of $E_t$ is detailed in Fig.~\ref{fig:extraction}.}
	\label{fig:quantity attention}
\end{figure}

The module takes as input the image features $I_g$, enhanced by the FeCom module, and the noise $z_t^4$ from the fourth block $B_4$ of the U-Net. It outputs perception information of quantity consistensy that can be interpreted by the U-Net to regulate the entire editing process. The module comprises three key components: the extraction module $E_t$, attention interaction, and U-Net injection. 
$E_t$ is designed to disentangle each object attribute from the whole $I_g$ while simultaneously capturing the global information of $I_g$. The attention interaction component translates the information extracted by $E_t$ into a format that is interpretable by the U-Net. Finally, the U-Net injection component leverages this transformed information to control the image editing process, preserving quantity consistency. Fig.~\ref{fig:quantity attention} presents the detailed structure of the QTTN module, with each step annotated to indicate its function. The notation “$\rightarrow$ (...)” represents the output dimensions at each stage.
\begin{figure}[htbp]
	\centering
	\begin{overpic}[width=0.48\textwidth, trim=18 10 18 15, clip]{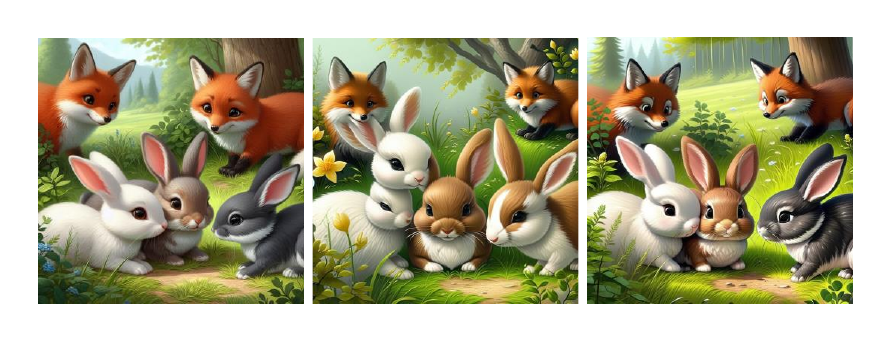}
		\put(9,-2){\small{\parbox{4cm}{\textcolor{black}{Reference}}}}
		\put(38.8,-2){\small{\parbox{4cm}{\textcolor{black}{w/o Extraction}}}}
		\put(73,-2){\small{\parbox{4cm}{\textcolor{black}{MoEdit (Ours)}}}}
	\end{overpic}
	\vspace{-2pt}
	\caption{The importance of the $E_t$. Please refer to Sec.~\ref{sec:Discussion of Extraction} for a comprehensive evaluation and detailed discussions regarding these configurations.}
	\label{fig:extraction}
\end{figure}
\subsubsection{Discussion of Extraction}
\label{sec:Discussion of Extraction}
In the QTTN module, $E_t$ is a critical component. Although implemented by a single FC layer, $E_t$ extracts clear and structured information about each object both individually and part of the whole $I_g$. This functionality cannot be directly incorporated into the attention interaction process. Fig.~\ref{fig:extraction} underscores the importance of $E_t$, with the “w/o Extraction” condition illustrating its absence.

Without $E_t$, the distinctive and separable object attributes in $I_g$ can still ensure quantity consistent perception. However, compared to MoEdit, the absence of $E_t$ leads to suboptimal object attributes extraction, reduced image details, and poor anatomical representations. This limitation arises because the Q, K, and V Projections within the attention interaction cannot effectively handle both the extraction of each object information and the transformation of information patterns simultaneously. Therefore, MoEdit incorporates an independent FC layer to ensure the high quality information of each object required for effective attention interactions. This unique functionality underscores the rationale for designating this FC layer as Extraction.

The quantitative comparisons are summarized in Table~\ref{tab:function of layernorm and extraction}. Notably, "w/o Extraction" exhibits a decline across multiple metrics, including image aesthetics, quality, text-image alignment, similarity between original and resulting images, and numerical accuracy.
\section{Supplementary Experiments}
\label{sec:Supplementary Experiments}
\begin{table*}[]
	\centering
	\renewcommand{\arraystretch}{1.2}
	\resizebox{\textwidth}{!}{%
		\begin{tabular}{lll}
			\hline
			\textbf{Method}         &   & \textbf{URL}                                                                   \\ \hline
			SSR-Encoder    & \cite{zhang2024ssr} & \url{https://github.com/Xiaojiu-z/SSR\_Encoder}                             \\
			$\lambda$-Eclipse     & \cite{patel2024lambda} & \url{https://github.com/eclipse-t2i/lambda-eclipse-inference}               \\
			IP-Adapter     & \cite{ye2023ip} & \url{https://github.com/tencent-ailab/IP-Adapter}                           \\
			Blip-diffusion & \cite{li2024blip} & \url{https://github.com/salesforce/LAVIS/tree/main/projects/blip-diffusion} \\
			MS-diffusion   & \cite{wang2024ms} & \url{https://github.com/MS-Diffusion/MS-Diffusion}                          \\
			Emu2           & \cite{sun2024generative} & \url{https://github.com/baaivision/Emu}                                     \\
			TurboEdit      & \cite{wu2024turboedit} & \url{https://betterze.github.io/TurboEdit/}                   \\ \hline\hline
			\textbf{Metric}         &   & \textbf{URL}                                                                   \\ \hline
			HyperIQA       & \cite{su2020blindly} & \url{https://github.com/SSL92/hyperIQA}                                     \\
			Q-Align        & \cite{wu2023q} & \url{https://github.com/Q-Future/Q-Align}                                   \\
			AesBench       & \cite{huang2024aesbench} & \url{https://github.com/yipoh/AesBench}                                     \\
			NIQE           & \cite{mittal2012making} & \url{https://github.com/csjunxu/Bovik\_NIQE\_SPL2013}                       \\
			CLIP Score     & \cite{hessel2021clipscore} & \url{https://github.com/Taited/clip-score}                                  \\
			LPIPS          & \cite{zhang2018unreasonable} & \url{https://github.com/richzhang/PerceptualSimilarity}                     \\ \hline
		\end{tabular}%
	}
	\caption{Code sources of seven comparison methods and six objective metrics}
	\label{tab:method and metric code sources}
\end{table*}
\subsection{Full Qualitative Comparisons}
\label{sec:Full Qualitative Comparisons}
In our main paper, MoEdit was comprehensively evaluated using six objective metrics, HyperIQA \cite{su2020blindly}, AesBench \cite{huang2024aesbench}, Q-Align \cite{wu2023q}, NIQE \cite{mittal2012making}, CLIP Score \cite{hessel2021clipscore}, and LPIPS \cite{zhang2018unreasonable}, and two subjective metrics, MOS and Numerical Accuracy. The evaluation compared MoEdit against seven methods: SSR-Encoder \cite{zhang2024ssr}, $\lambda$-Eclipse \cite{patel2024lambda}, IP-Adapter \cite{ye2023ip}, Blip-diffusion \cite{li2024blip}, MS-diffusion \cite{wang2024ms}, Emu2 \cite{sun2024generative}, and TurboEdit \cite{wu2024turboedit}. This section provides additional details on these comparison methods and summarizes the code sources for all methods and evaluation metrics in Table~\ref{tab:method and metric code sources}, facilitating future research replication.\\
\noindent\textbf{SSR-Encoder.} This method is tailored for subject-driven generative tasks, encoding selective subject representations. By focusing on extracting and representing subject-specific features, it enhances the ability to maintain subject consistency in generated outputs.\\
\noindent\textbf{$\lambda$-Eclipse.} A multi-concept text-to-image generation model that leverages the latent space of CLIP to achieve efficient multimodal alignment. This method allows the flexible integration of multiple concepts while preserving semantic consistency between textual and visual inputs.\\
\noindent\textbf{IP-Adapter.} A text-compatible image-prompt adapter designed for text-to-image diffusion models. The introduction of efficient adapter modules enables the model to utilize image prompts to enhance generation quality while maintaining compatibility with textual input.\\
\noindent\textbf{Blip-diffusion.} This method combines pre-trained subject representations with text-to-image generation and editing. By integrating precise subject modeling, it offers more controllable image generation and editing while improving semantic alignment between text and images.\\
\noindent\textbf{MS-diffusion.}~A method for multi-subject image personalization that integrates layout guidance for structured scene control.~This method generates multi-subject, layout-accurate images in zero-shot scenarios, enhancing the diversity and consistency of outputs.\\
\noindent\textbf{Emu2.} This study investigates the contextual learning capabilities of generative multimodal models, showcasing their ability to perform diverse tasks in unsupervised settings via input prompts. The findings highlight the potential of these models for cross-task generalization and implicit learning, underscoring their flexibility and adaptability.\\
\noindent\textbf{TurboEdit.} An efficient text-driven image editing method that rapidly generates modifications aligned with textual descriptions. By optimizing the editing workflow, it provides a real-time, interactive image editing experience.

\begin{figure*}[htbp]
	\centering
	\begin{overpic}[width=0.98\textwidth, trim=18 10 18 15, clip]{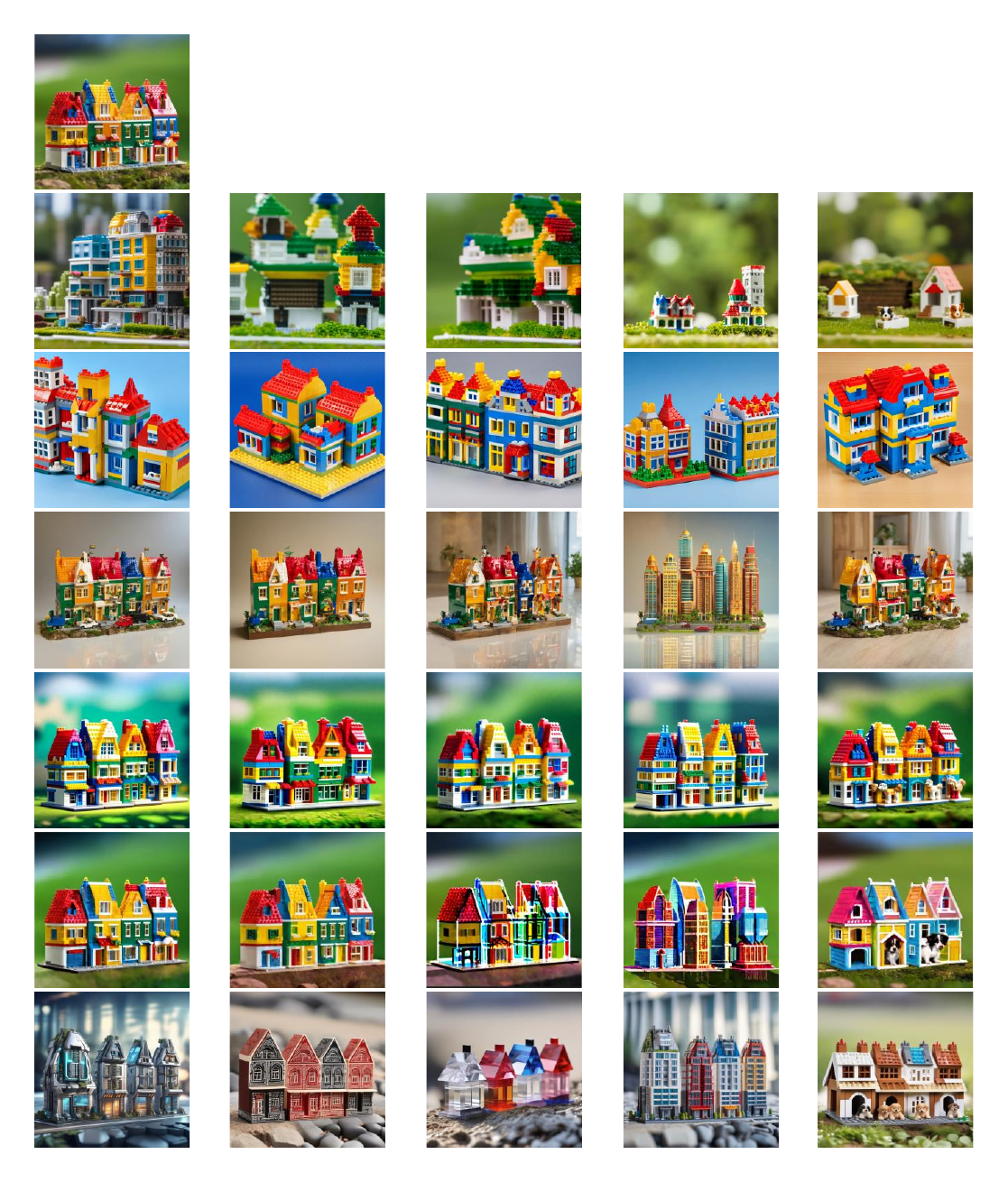}
		\put(35,92){{\LARGE\parbox{4cm}{\textcolor{lightblue}{\textit{four lego houses}}}}}
		\put(-1.5,88.5){\rotatebox{90}{\parbox{4cm}{{Reference}}}}
		\put(-1.5,74.5){\rotatebox{90}{\parbox{4cm}{{IP-Adapter}}}}
		\put(-1.5,59){\rotatebox{90}{\parbox{4cm}{{Blip-diffusion}}}}
		\put(-1.5,45){\rotatebox{90}{\parbox{4cm}{{MS-diffusion}}}}
		\put(-1.5,34){\rotatebox{90}{\parbox{4cm}{{Emu2}}}}
		\put(-1.5,18){\rotatebox{90}{\parbox{4cm}{{TurboEdit}}}}
		\put(-1.5,2.5){\rotatebox{90}{\parbox{4cm}{{MoEdit (Ours)}}}}
		\put(-0.5,-1.5){{\small\parbox{3cm}{\centering{\textit{“...futuristic metropolis style”}}}}}
		\put(17.5,-1.5){{\small\parbox{3cm}{{\textit{“...block print style”}}}}}
		\put(35.4,-1.5){{\small\parbox{3cm}{{\textit{“...made of crystal”}}}}}
		\put(53.4,-1.5){{\small\parbox{3cm}{{\textit{“\textcolor{red}{$\rightarrow$}skyscrapers”}}}}}
		\put(70,-1.5){{\small\parbox{3cm}{{\textit{“\textcolor{red}{$\rightarrow$}puppy houses”}}}}}
	\end{overpic}
	\vspace{10pt}
	\caption{Qualitative comparisons among a set of four objects.}
	\label{fig:four-comparisons}
\end{figure*}

\begin{figure*}[htbp]
	\centering
	\begin{overpic}[width=0.98\textwidth, trim=18 10 18 15, clip]{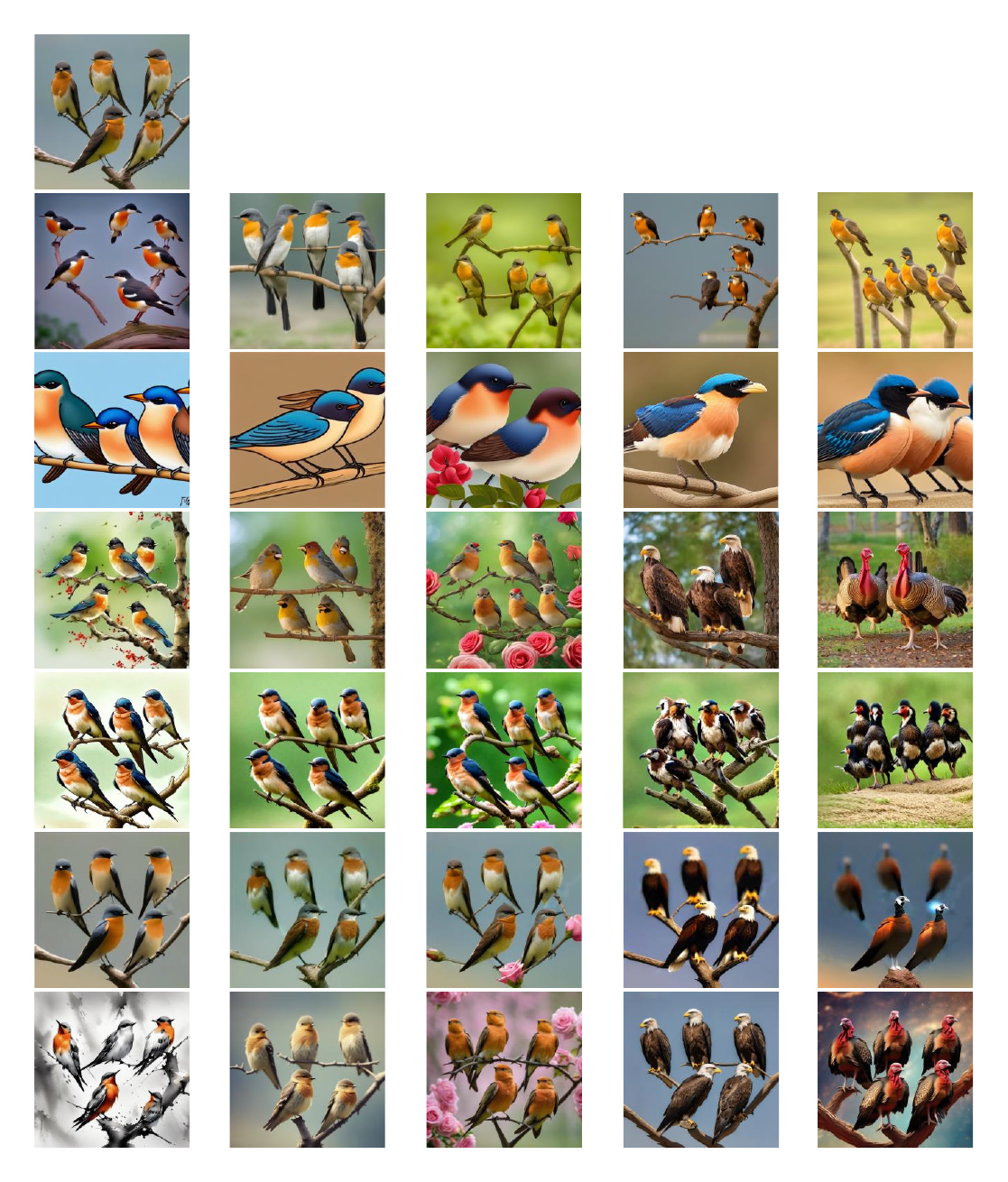}
		\put(42,92){{\LARGE\parbox{4cm}{\textcolor{lightblue}{\textit{five birds}}}}}
		\put(-1.5,88.5){\rotatebox{90}{\parbox{4cm}{{Reference}}}}
		\put(-1.5,74.5){\rotatebox{90}{\parbox{4cm}{{IP-Adapter}}}}
		\put(-1.5,59){\rotatebox{90}{\parbox{4cm}{{Blip-diffusion}}}}
		\put(-1.5,45.5){\rotatebox{90}{\parbox{4cm}{{MS-diffusion}}}}
		\put(-1.5,34){\rotatebox{90}{\parbox{4cm}{{Emu2}}}}
		\put(-1.5,18){\rotatebox{90}{\parbox{4cm}{{TurboEdit}}}}
		\put(-1.5,3){\rotatebox{90}{\parbox{4cm}{{MoEdit (Ours)}}}}
		\put(0.3,-1.5){{\small\parbox{3cm}{{\textit{“...ink painting style”}}}}}
		\put(18.3,-1.5){{\small\parbox{3cm}{{\textit{“...faded film style”}}}}}
		\put(36.5,-1.5){{\small\parbox{2cm}{\centering{\textit{“...with roses in background”}}}}}
		\put(55.4,-1.5){{\small\parbox{3cm}{{\textit{“\textcolor{red}{$\rightarrow$}eagles”}}}}}
		\put(69.,-1.5){{\small\parbox{3cm}{\centering{\textit{“\textcolor{red}{$\rightarrow$}turkeys, with universe background”}}}}}
	\end{overpic}
	\vspace{10pt}
	\caption{Qualitative comparisons among a set of five objects.}
	\label{fig:five-comparisons}
\end{figure*}

\begin{figure*}[htbp]
	\centering
	\begin{overpic}[width=0.98\textwidth, trim=18 10 18 15, clip]{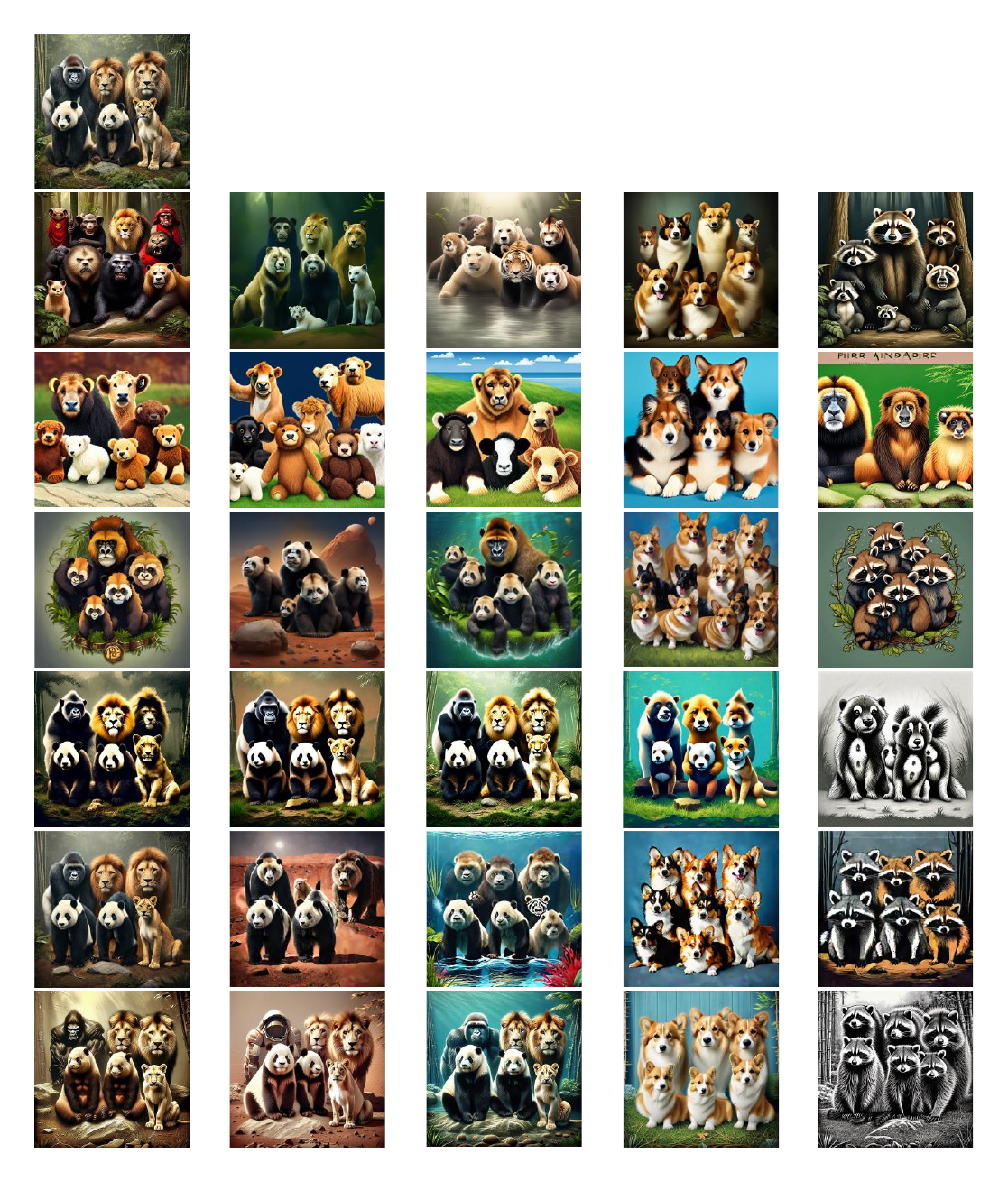}
		\put(26,92){{\LARGE\parbox{10cm}{\textcolor{lightblue}{\textit{three lions, two pandas and a gorilla}}}}}
		\put(-1.5,88.5){\rotatebox{90}{\parbox{4cm}{{Reference}}}}
		\put(-1.5,74.5){\rotatebox{90}{\parbox{4cm}{{IP-Adapter}}}}
		\put(-1.5,59){\rotatebox{90}{\parbox{4cm}{{Blip-diffusion}}}}
		\put(-1.5,45.5){\rotatebox{90}{\parbox{4cm}{{MS-diffusion}}}}
		\put(-1.5,34){\rotatebox{90}{\parbox{4cm}{{Emu2}}}}
		\put(-1.5,18){\rotatebox{90}{\parbox{4cm}{{TurboEdit}}}}
		\put(-1.5,3){\rotatebox{90}{\parbox{4cm}{{MoEdit (Ours)}}}}
		\put(2,-1.5){{\small\parbox{3cm}{{\textit{“...diablo style”}}}}}
		\put(19,-1.5){{\small\parbox{3cm}{{\textit{“...on the mars”}}}}}
		\put(36.8,-1.5){{\small\parbox{2cm}{{\textit{“...in the sea”}}}}}
		\put(52.5,-1.5){{\small\parbox{2.6cm}{\centering{\textit{“\textcolor{red}{$\rightarrow$}corgis, with blue background”}}}}}
		\put(70.3,-1.5){{\small\parbox{2.3cm}{\centering{\textit{“\textcolor{red}{$\rightarrow$}raccoons, line art style”}}}}}
	\end{overpic}
	\vspace{10pt}
	\caption{Qualitative comparisons among a set of six objects.}
	\label{fig:six-comparisons}
\end{figure*}

\begin{figure*}[htbp]
	\centering
	\begin{overpic}[width=0.98\textwidth, trim=18 10 18 15, clip]{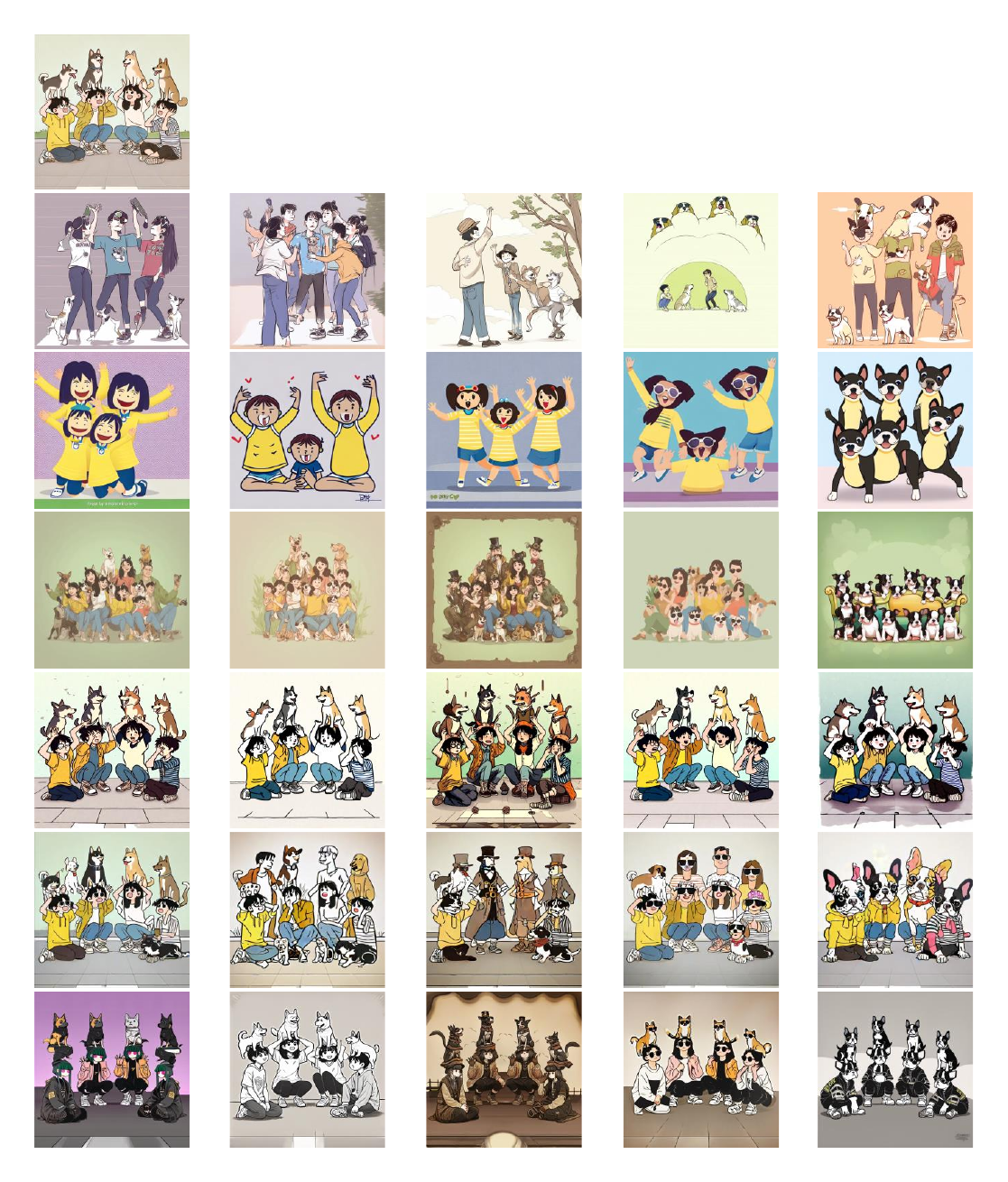}
		\put(30,92){{\LARGE\parbox{10cm}{\textcolor{lightblue}{\textit{four people and four dogs}}}}}
		\put(-1.5,88.5){\rotatebox{90}{\parbox{4cm}{{Reference}}}}
		\put(-1.5,74.5){\rotatebox{90}{\parbox{4cm}{{IP-Adapter}}}}
		\put(-1.5,59){\rotatebox{90}{\parbox{4cm}{{Blip-diffusion}}}}
		\put(-1.5,45.5){\rotatebox{90}{\parbox{4cm}{{MS-diffusion}}}}
		\put(-1.5,34){\rotatebox{90}{\parbox{4cm}{{Emu2}}}}
		\put(-1.5,18){\rotatebox{90}{\parbox{4cm}{{TurboEdit}}}}
		\put(-1.5,3){\rotatebox{90}{\parbox{4cm}{{MoEdit (Ours)}}}}
		\put(0.5,-1.5){{\small\parbox{3cm}{{\textit{“...cyberpunck style”}}}}}
		\put(19,-1.5){{\small\parbox{3cm}{{\textit{“...sketch style”}}}}}
		\put(34.7,-1.5){{\small\parbox{3cm}{{\textit{“...steampunck style”}}}}}
		\put(52.7,-1.5){{\small\parbox{2.6cm}{{\textit{“...with sunglasses”}}}}}
		\put(67.8,-1.5){{\small\parbox{3.4cm}{\centering{\textit{“\textcolor{red}{$\rightarrow$}boston terrier puppies, cyberpunck style”}}}}}
	\end{overpic}
	\vspace{10pt}
	\caption{Qualitative comparisons among a set of eight objects.}
	\label{fig:eight-comparisons}
\end{figure*}

\begin{figure*}[htbp]
	\centering
	\begin{overpic}[width=0.98\textwidth, trim=18 10 18 15, clip]{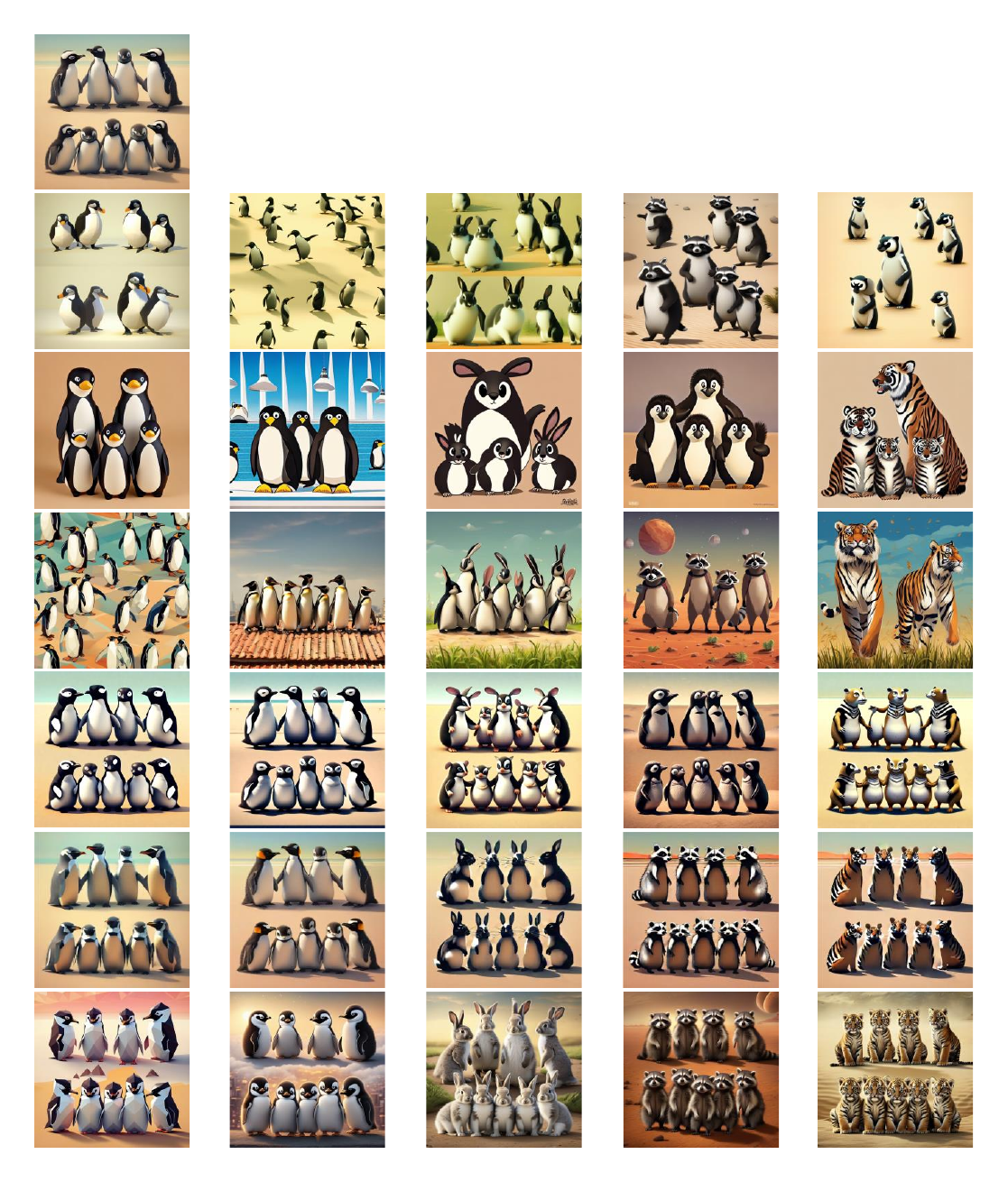}
		\put(38,92){{\LARGE\parbox{10cm}{\textcolor{lightblue}{\textit{nine penguins}}}}}
		\put(-1.5,88.5){\rotatebox{90}{\parbox{4cm}{{Reference}}}}
		\put(-1.5,74.5){\rotatebox{90}{\parbox{4cm}{{IP-Adapter}}}}
		\put(-1.5,59){\rotatebox{90}{\parbox{4cm}{{Blip-diffusion}}}}
		\put(-1.5,45.5){\rotatebox{90}{\parbox{4cm}{{MS-diffusion}}}}
		\put(-1.5,34){\rotatebox{90}{\parbox{4cm}{{Emu2}}}}
		\put(-1.5,18){\rotatebox{90}{\parbox{4cm}{{TurboEdit}}}}
		\put(-1.5,3){\rotatebox{90}{\parbox{4cm}{{MoEdit (Ours)}}}}
		\put(1.3,-1.5){{\small\parbox{3cm}{{\textit{“...low poly style”}}}}}
		\put(18.5,-1.5){{\small\parbox{3cm}{{\textit{“...on the rooftop”}}}}}
		\put(34.6,-1.5){{\small\parbox{2.8cm}{\centering{\textit{“\textcolor{red}{$\rightarrow$}rabbits, in a natural field”}}}}}
		\put(52.5,-1.5){{\small\parbox{2.6cm}{\centering{\textit{“\textcolor{red}{$\rightarrow$}raccoons, on the mars”}}}}}
		\put(68.7,-1.5){{\small\parbox{3cm}{\centering{\textit{“\textcolor{red}{$\rightarrow$}tigers, in the desert”}}}}}
	\end{overpic}
	\vspace{10pt}
	\caption{Qualitative comparisons among a set of nine objects.}
	\label{fig:nine-comparisons}
\end{figure*}

\begin{figure*}[htbp]
	\centering
	\begin{overpic}[width=0.98\textwidth, trim=18 10 18 15, clip]{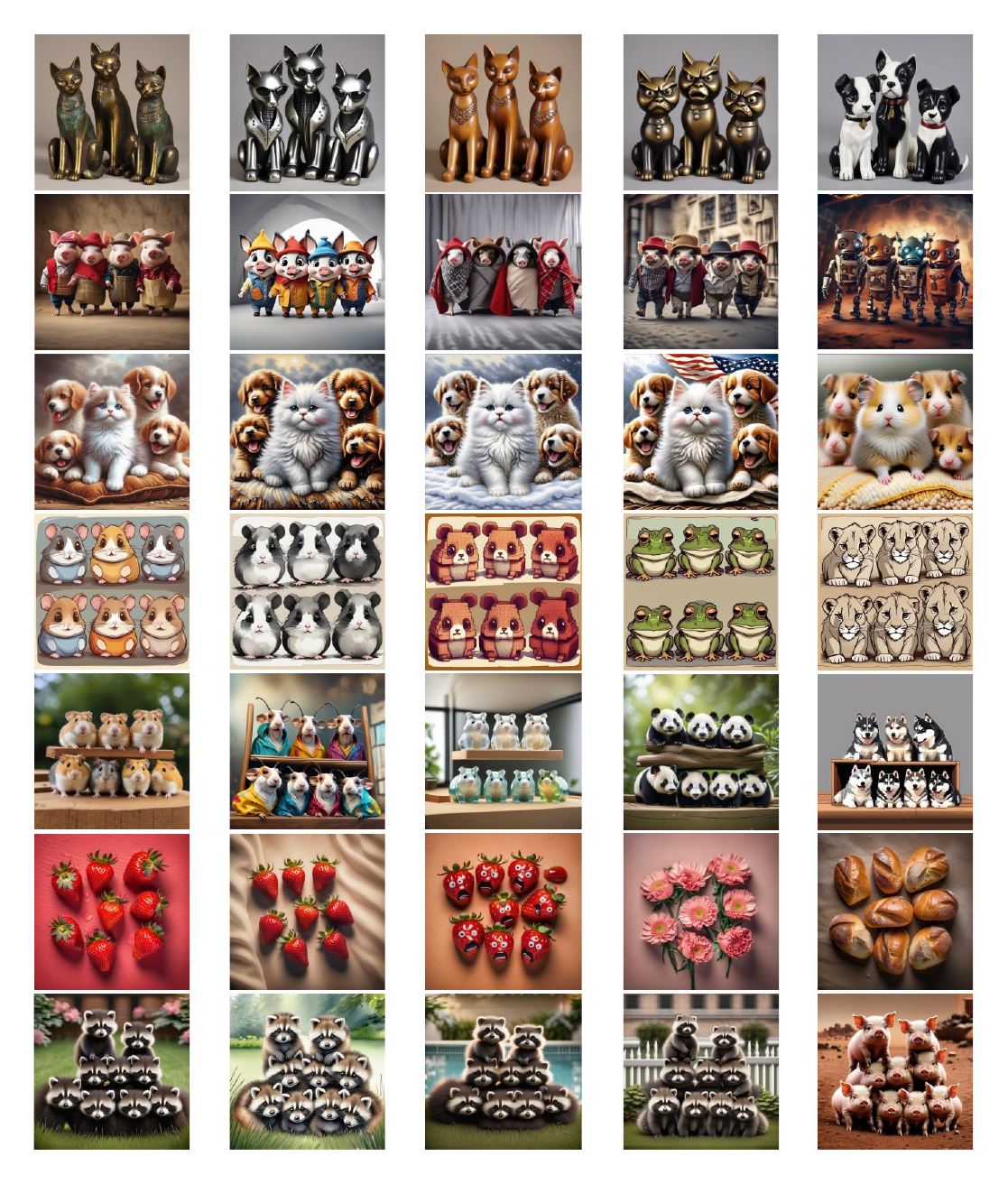}
		\put(3.5,100){{\parbox{4cm}{{Reference}}}}
		\put(47.5,100){{\parbox{4cm}{{Outputs}}}}
		\put(14.7,87.2){{\small\rotatebox{90}{\parbox{10cm}{\textcolor{lightblue}{\textit{three bronze cats}}}}}}	
		\put(31.3,85.5){{\footnotesize\rotatebox{90}{\parbox{3cm}{\centering{\textit{“...with sunglasses, futuristic urban style”}}}}}}
		\put(49.5,85.2){{\footnotesize\rotatebox{90}{\parbox{3cm}{\centering{\textit{“...made of wood”}}}}}}
		\put(66.8,85.5){{\footnotesize\rotatebox{90}{\parbox{3cm}{\centering{\textit{“...with smilling faces”}}}}}}
		\put(83.1,85.4){{\footnotesize\rotatebox{90}{\parbox{3cm}{\centering{\textit{“\textcolor{red}{$\rightarrow$}black and white puppies”}}}}}}
		\put(14.7,74){{\small\rotatebox{90}{\parbox{10cm}{\textcolor{lightblue}{\textit{four little pigs}}}}}}	
		\put(32,71.2){{\footnotesize\rotatebox{90}{\parbox{3cm}{\centering{\textit{“...cartoon style”}}}}}}
		\put(49.5,71.2){{\footnotesize\rotatebox{90}{\parbox{3cm}{\centering{\textit{“...with scarves”}}}}}}
		\put(66.8,71.5){{\footnotesize\rotatebox{90}{\parbox{3cm}{\centering{\textit{“...in the street”}}}}}}
		\put(84,71.3){{\footnotesize\rotatebox{90}{\parbox{3cm}{\centering{\textit{“\textcolor{red}{$\rightarrow$}robots, on the mars”}}}}}}
		\put(14.7,57.7){{\small\rotatebox{90}{\parbox{10cm}{\textcolor{lightblue}{\textit{four puppies and a cat}}}}}}	
		\put(31.4,57){{\footnotesize\rotatebox{90}{\parbox{3cm}{\centering{\textit{“...richly textured oil painting”}}}}}}
		\put(49.5,57.2){{\footnotesize\rotatebox{90}{\parbox{3cm}{\centering{\textit{“...in the snow”}}}}}}
		\put(65.9,57.1){{\footnotesize\rotatebox{90}{\parbox{3cm}{\centering{\textit{“...with flag in background”}}}}}}
		\put(84,57.2){{\footnotesize\rotatebox{90}{\parbox{3cm}{\centering{\textit{“\textcolor{red}{$\rightarrow$}mice”}}}}}}
		\put(14.7,46.5){{\small\rotatebox{90}{\parbox{10cm}{\textcolor{lightblue}{\textit{six hamsters}}}}}}	
		\put(32,42.8){{\footnotesize\rotatebox{90}{\parbox{3cm}{\centering{\textit{“...ink painting style”}}}}}}
		\put(49.5,42.8){{\footnotesize\rotatebox{90}{\parbox{3cm}{\centering{\textit{“...voxel style”}}}}}}
		\put(66.8,43){{\footnotesize\rotatebox{90}{\parbox{3cm}{\centering{\textit{“\textcolor{red}{$\rightarrow$}frogs”}}}}}}
		\put(84,43){{\footnotesize\rotatebox{90}{\parbox{3cm}{\centering{\textit{“\textcolor{red}{$\rightarrow$}lions”}}}}}}
		\put(14.7,33){{\small\rotatebox{90}{\parbox{10cm}{\textcolor{lightblue}{\textit{seven mice}}}}}}	
		\put(31.3,27.8){{\footnotesize\rotatebox{90}{\parbox{3.3cm}{\centering{\textit{“...vibrant portrait painting of Salvador Dalí”}}}}}}
		\put(49.5,28.8){{\footnotesize\rotatebox{90}{\parbox{3cm}{\centering{\textit{“...made of glass”}}}}}}
		\put(66.8,28.8){{\footnotesize\rotatebox{90}{\parbox{3cm}{\centering{\textit{“\textcolor{red}{$\rightarrow$}pandas”}}}}}}
		\put(83.1,28.8){{\footnotesize\rotatebox{90}{\parbox{3cm}{\centering{\textit{“\textcolor{red}{$\rightarrow$}huskies, pixel art style”}}}}}}
		\put(14.7,16.5){{\small\rotatebox{90}{\parbox{10cm}{\textcolor{lightblue}{\textit{eight strawberries}}}}}}	
		\put(32,14){{\footnotesize\rotatebox{90}{\parbox{3.3cm}{\centering{\textit{“...cinematic shot”}}}}}}
		\put(49.5,14.8){{\footnotesize\rotatebox{90}{\parbox{3cm}{\centering{\textit{“...frightened faces”}}}}}}
		\put(66.8,14.8){{\footnotesize\rotatebox{90}{\parbox{3cm}{\centering{\textit{“\textcolor{red}{$\rightarrow$}flowers”}}}}}}
		\put(84,15.2){{\footnotesize\rotatebox{90}{\parbox{3cm}{\centering{\textit{“\textcolor{red}{$\rightarrow$}breads”}}}}}}
		\put(14.7,4){{\small\rotatebox{90}{\parbox{10cm}{\textcolor{lightblue}{\textit{nine raccoons}}}}}}	
		\put(32,0){{\footnotesize\rotatebox{90}{\parbox{3.3cm}{\centering{\textit{“...watercolor style”}}}}}}
		\put(49.5,0.8){{\footnotesize\rotatebox{90}{\parbox{3cm}{\centering{\textit{“...by the pool”}}}}}}
		\put(66,0.3){{\footnotesize\rotatebox{90}{\parbox{3cm}{\centering{\textit{“...with building in background”}}}}}}
		\put(84,0.7){{\footnotesize\rotatebox{90}{\parbox{3cm}{\centering{\textit{“\textcolor{red}{$\rightarrow$}pigs, on the mars”}}}}}}
	\end{overpic}
	\vspace{10pt}
	\caption{Additional qualitative results.}
	\label{fig:qualitative results}
\end{figure*}

Due to space constraints in the main paper, qualitative comparisons were limited to conditions 3, 7, and 9+. In this section, results for the remaining conditions are presented in Figs.~\ref{fig:four-comparisons}–\ref{fig:nine-comparisons}. Furthermore, Fig.~\ref{fig:qualitative results} offers additional visual demonstrations of MoEdit output.
\subsection{Visualization of Ablation Results}
\label{sec:scale}
\begin{figure*}[htbp]
	\centering
	\begin{overpic}[width=0.98\textwidth, trim=18 10 18 15, clip]{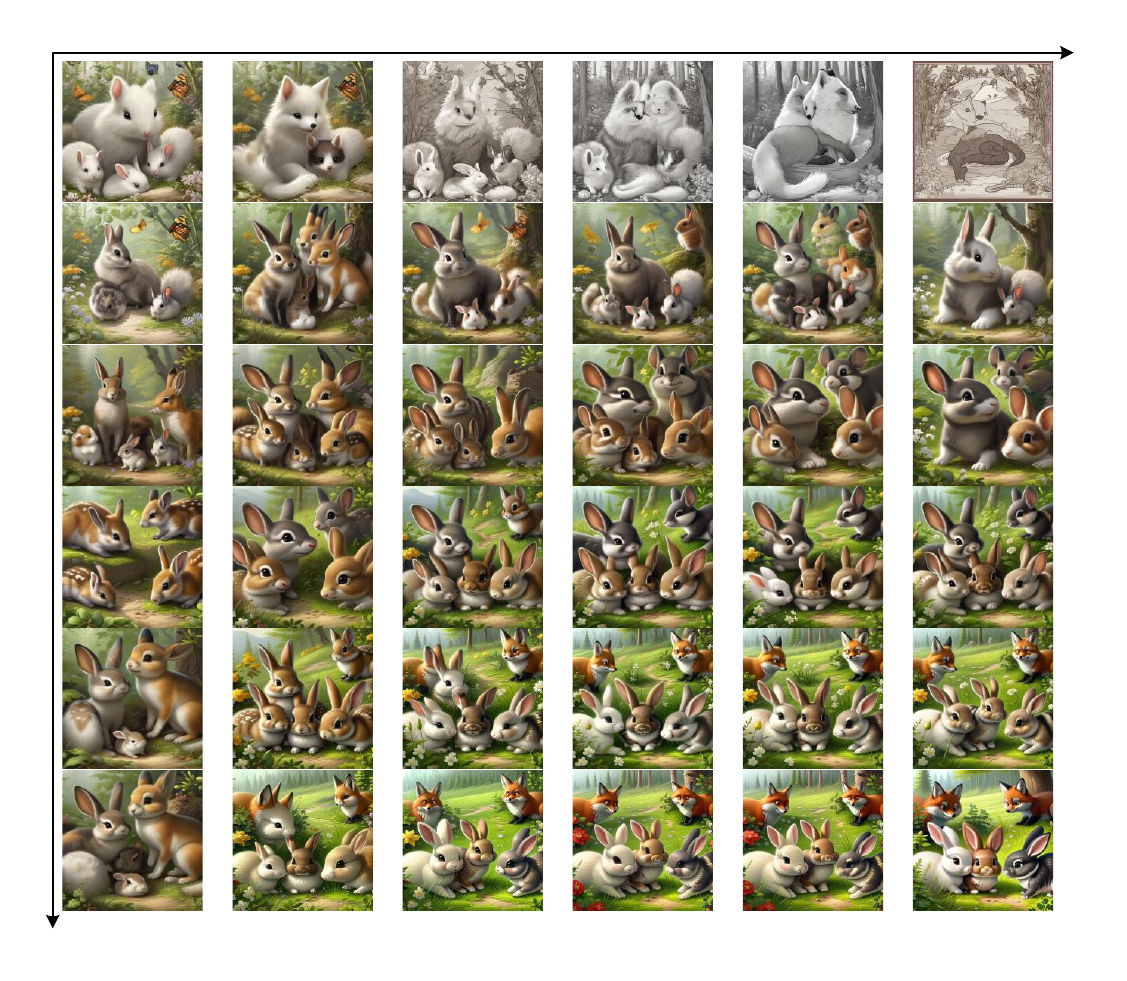}
		\put(3.5,87){{\parbox{4cm}{{($\lambda$ = 0, $\beta$ = 0)}}}}
		\put(23,87){{\parbox{4cm}{{\textcolor{red}{$\uparrow$} 0.2}}}}
		\put(39,87){{\parbox{4cm}{{\textcolor{red}{$\uparrow$} 0.4}}}}
		\put(55,87){{\parbox{4cm}{{\textcolor{red}{$\uparrow$} 0.6}}}}
		\put(71,87){{\parbox{4cm}{{\textcolor{red}{$\uparrow$} 0.8}}}}
		\put(87.5,87){{\parbox{4cm}{{\textcolor{red}{$\uparrow$} 1.0}}}}
		\put(99,86){{\parbox{4cm}{{$\lambda$}}}}
		\put(1,0.6){{\parbox{4cm}{{$\beta$}}}}
		\put(-0.8,73){{\rotatebox{90}{\parbox{4cm}{{($\lambda$ = 0, $\beta$ = 0)}}}}}
		\put(-0.8,63){{\rotatebox{90}{\parbox{4cm}{\textcolor{blue}{$\uparrow$} 0.2}}}}
		\put(-0.8,49){{\rotatebox{90}{\parbox{4cm}{\textcolor{blue}{$\uparrow$} 0.4}}}}
		\put(-0.8,36){{\rotatebox{90}{\parbox{4cm}{\textcolor{blue}{$\uparrow$} 0.6}}}}
		\put(-0.8,22.5){{\rotatebox{90}{\parbox{4cm}{\textcolor{blue}{$\uparrow$} 0.8}}}}
		\put(-0.8,8.5){{\rotatebox{90}{\parbox{4cm}{\textcolor{blue}{$\uparrow$} 1.0}}}}
	\end{overpic}
	\vspace{-8pt}
	\caption{Scale variation. The reference image of all results are refered to Fig.~\ref{fig:layernorm}. The top-left image illustrates the outcome when ($\lambda$ = 0, $\beta$ = 0). Each row corresponds to results obtained with a fixed $\beta$ and varying $\lambda$. From left to right, $\lambda$ increases incrementally, as indicated by \textcolor{red}{$\uparrow$} $x$, where $x$ denotes the increment from the baseline $\lambda$ = 0. Similarly, each column corresponds to results generated with a fixed $\lambda$ and varying $\beta$. From top to bottom, $\beta$ increases incrementally, marked by \textcolor{blue}{$\uparrow$} $x$, where $x$ represents the increment from the baseline $\beta$ = 0. Detailed discussions of these results are provided in Sec.~\ref{sec:scale}.}
	\label{fig:scale_variation}
\end{figure*}

\noindent\textbf{Scale variation.} Due to the limited space in the main paper, we presented only the results for $\lambda$ = 1 with varying $\beta$ and $\beta$ = 1 with varying $\lambda$. However, these results alone are insufficient to fully capture the mechanisms of the two modules. To address this, more comprehensive results are provided in Fig.~\ref{fig:scale_variation}, offering a clearer demonstration of the interaction between the modules.

When $\lambda$ is fixed, increasing $\beta$ enhances the ability of the QTTN module to disentangle each object attribute from the whole image and to extract global information, resulting in higher-quality outputs. This improvement, however, relies heavily on the input features $I_g$ of the QTTN module, which must exhibit sufficient distinction and separability of object attributes. Conversely, when $\beta$ is fixed, reducing $\lambda$ increases the degree of aliasing of object attributes. Consequently, the QTTN module becomes less effective at extracting information of each object both individual and part of the whole, leading to a deterioration in the output quality.
\subsection{Limitations}
\label{sec:limitations}
\begin{figure*}[htbp]
	\centering
	\begin{overpic}[width=1.0\textwidth, trim=18 10 18 15, clip]{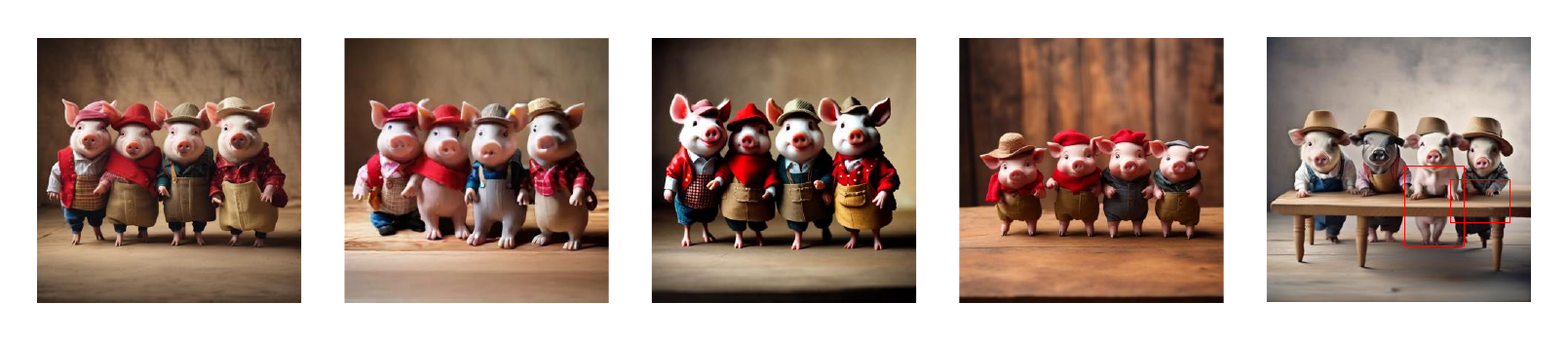}
		\put(4.5,-0.5){{\parbox{4cm}{{Reference}}}}
		\put(25.5,-0.5){{\small\parbox{4cm}{{TurboEdit}}}}
		\put(48,-0.5){{\small\parbox{4cm}{{Emu2}}}}
		\put(65.8,-0.5){{\small\parbox{4cm}{{MS-diffusion}}}}
		\put(86,-0.5){{\small\parbox{4cm}{{MoEdit (Ours)}}}}
	\end{overpic}
	\caption{The illustration of limitations. The reference refers to the input image. All output images are edited based on the text prompts \textit{“on the table”}. For more discussion, please refer to Sec.~\ref{sec:limitations}.}
	\label{fig:limitation}
\end{figure*}

When the spatial relationship between objects and their surrounding 3D environment changes, MoEdit often struggles to deliver satisfactory editing results. This limitation stems from the baseline model, SDXL \cite{podell2023sdxl}, which lacks the capability to effectively construct accurate 3D relationships in multi-object scenes. As illustrated in Fig.~\ref{fig:limitation}, prompts like \textit{“on the table”} require consideration of not only the interaction between existing objects and their environment but also the 3D spatial alignment between the newly introduced table and multiple objects. This challenge results in MoEdit producing artifacts such as misaligned object segments, truncation, and loss of surrounding elements like clothing. Meanwhile, other methods exhibit even poorer performance, either failing to achieve meaningful edits or simply modifying the ground to a wooden texture, thereby creating a illusion of objects being placed on a table. To address these challenges, future work could focus on integrating 3D environmental data, transcending the limitations of 2D information, to achieve better alignment and coherence between multi-object and their scenes.


\end{document}